\documentclass[pdflatex,sn-mathphys-num]{sn-jnl}

\usepackage{graphicx}%
\usepackage{multirow}%
\usepackage{amsmath,amssymb,amsfonts}%
\usepackage{amsthm}%
\usepackage{mathrsfs}%
\usepackage[title]{appendix}%
\usepackage{xcolor}%
\usepackage{textcomp}%
\usepackage{manyfoot}%
\usepackage{booktabs}%
\usepackage{algorithm}%
\usepackage{algorithmicx}%
\usepackage{algpseudocode}%
\usepackage{listings}%
\usepackage{caption}
\usepackage{multirow}
\theoremstyle{thmstyleone}%
\theoremstyle{thmstyletwo}%

\theoremstyle{thmstylethree}%

\begin{document}

\title[Article Title]{AI-Driven Diabetic Retinopathy Diagnosis Enhancement through Image Processing and Salp Swarm Algorithm-Optimized Ensemble Network}


\author*[1,2]{\fnm{Saif Ur Rehman} \sur{Khan}}\email{saifurrehman.khan@csu.edu.cn}

\author*[3,4]{\fnm{Muhammad Nabeel} \sur{Asim}}\email{muhammad\_nabeel.asim@dfki.de}
\author[3,4]{\fnm{Sebastian} \sur{ Vollmer}}\email{sebastian.vollmer@dfki.de}
\author[3,4,5]{\fnm{Andreas} \sur{Dengel}}\email{andreas.dengel@dfki.de}

\affil*[1]{\orgdiv{School of Computer Science \& Engineering}, \orgname{Central South University}, \orgaddress{\city{Changsha}, \country{China}}}

\affil*[2]{\orgdiv{Department of Computer Science}, \orgname{National University of Computer \& Emerging Science}, \orgaddress{\city{Islamabad}, \country{Pakistan}}}

\affil[3]{\orgdiv{German Research Center for Artificial Intelligence}, \orgaddress{ \city{Kaiserslautern}, \postcode{67663}, \country{Germany}}}
\affil[4]{\orgdiv{Intelligentx GmbH (intelligentx.com)}, \orgaddress{ \city{Kaiserslautern}, \country{Germany}}}
\affil[5]{\orgdiv{Department of Computer Science}, \orgname{Rhineland-Palatinate Technical University of Kaiserslautern-Landau} \orgaddress{ \city{Kaiserslautern}, \postcode{67663}, \country{Germany}}}

\abstract{Diabetic retinopathy is a leading cause of blindness in diabetic patients and early detection plays a crucial role in preventing vision loss. Traditional diagnostic methods are often time-consuming and prone to errors. The emergence of deep learning techniques has provided innovative solutions to improve diagnostic efficiency. However, single deep learning models frequently face issues related to extracting key features from complex retinal images. To handle this problem, we present an effective ensemble method for DR diagnosis comprising four main phases: image pre-processing, selection of backbone pre-trained models, feature enhancement, and optimization. Our methodology initiates with the pre-processing phase, where we apply CLAHE to enhance image contrast and Gamma correction is then used to adjust the brightness for better feature recognition. We then apply Discrete Wavelet Transform (DWT) for image fusion by combining multi-resolution details to create a richer dataset. Then, we selected three pre-trained models with the best performance named DenseNet169, MobileNetV1, and Xception for diverse feature extraction. To further improve feature extraction, an improved residual block is integrated into each model. Finally, the predictions from these base models are then aggregated using weighted ensemble approach, with the weights optimized by using Salp Swarm Algorithm (SSA).SSA intelligently explores the weight space and finds the optimal configuration of base architectures to maximize the performance of the ensemble model. The proposed model is evaluated on the multiclass Kaggle APTOS 2019 dataset and obtained 88.52\% accuracy. The performance of the proposed model is further validated using advanced evaluation metrics including confusion matrix (CM), receiver operating characteristic (ROC) curve, Precision-Recall curve (P-R curve) visualizations. These results demonstrate the robustness and superior performance of our optimized ensemble model in DR detection.}



\maketitle

\section{Introduction}\label{sec1}

Diabetic retinopathy (DR) arises when high blood sugar levels harm the blood vessels in the retina, which is responsible for vision. It often starts without symptoms but can progress to blurred vision, dark spots by resulting in vision loss. It is responsible for 12\% of new blindness cases and is the leading cause of blindness in individuals aged 20 to 64 in the developed country \cite{tamkin2013deep}. It has influenced 75 percent of those with diabetes over 15 years. Chronic hyperglycemia causes retinal microvascular and neurovascular damage by indicating to complications like proliferative diabetic retinopathy (PDR) and diabetic macular edema (DME) \cite{yau2012global}. DR is a major cause of blindness globally by affecting over 158.2 million people and this number is projected to exceed 191 million by 2030 \cite{jabbar2024lesion}. The rise in diabetes cases is directly linked to an increase in vision loss. In 2022, approximately 9\% of the global population is affected by diabetes, and this number is expected to grow to approximately 12\% by 2030 \cite{senapati2024artificial}. DR is the leading cause of vision loss in individuals with diabetes if it is not treated early. The condition can also impact multiple organs like heart, kidneys, and eyes. Early screening is essential to prevent DR and early detection can also aid to save up to 90 percent of diabetic patients \cite{vashist2011role}. DR is categorized into 2 main types, which are proliferative DR and non-proliferative DR \cite{mateen2020automatic} \cite{alyoubi2020diabetic} . Non-proliferative DR is further allotted into three stages, such as mild, moderate, and severe \cite{bhardwaj2021hierarchical}.

The manual evaluation of retinal images for detecting DR by ophthalmologists is time-consuming and costly that causes a risk of errors \cite{vij2023systematic}. Deep learning (DL) \cite{khan2025detection} methods have achieved significance in medical applications due to their ability to excel in the classification process while also enhancing accuracy. Deep Convolutional neural networks (DCNNs) are effective for medical image analysis, as they can automatically extract relevant features from various types of medical imaging modalities \cite{hekmat2025attention,khan2024deep,khan2025optimized}. These capabilities have led researchers to develop a variety of CNN-based techniques for improving diagnostic performance automatically in recent years. However, CNNs can struggle with retinal images that lack sufficient clarity, or they consist of noise, which can hinder accurate detection. To address these limitations, image fusion techniques\cite{bibi2020automated,ur2018diabetic} play a critical role in diabetic retinopathy by combining high-quality enhanced images, thereby improving overall image clarity and emphasizing key features.

In medical image classification, such as diagnosing DR abnormalities, individual models often face significant challenges. These models may struggle with the complexity and variability of analyzing detailed medical datasets, including noise, imbalance, and the diversity of patient demographics \cite{patel2020transfer}. A single model might not be able to process such diverse data effectively by yielding lower accuracy performance. To tackle these issues, ensemble learning techniques were introduced to enhance generalization and reduce the risk of overfitting \cite{saleh2018learning,vij2024hybrid}. These methods combine multiple models to improve predictive accuracy and robustness. By aggregating predictions from several models, ensemble learning provides more reliable and accurate diagnoses by leading to improved survival rates of patients. Despite their advantages, ensemble models can also face complications. If the individual models in the ensemble are highly correlated, they may cause overfitting. Furthermore, optimizing the integration of models and adjusting their weights can be computationally cost effective due to hit and trial methods. Conventional approaches may struggle to search for the extensive range of parameters effectively. To tackle these issues, metaheuristic-based optimization techniques \cite{bilal2025amalgamation,beevi2023multi} have been extensively used to optimize the ensemble process and improve overall performance in the classification of various diseases, including DR and others medical disease. These optimization methods aid enhanced model performance by considering them valuable tools in medical imaging related tasks.
\subsection{Novelty and Motivation}
DR is one of the key problems of blindness among diabetic patients by making early detections crucial for preventing severe vision loss. Timely diagnosis significantly improves patient outcomes and reduces the risks of permanent damage. This study was motivated by the need to enhance the classification of DR via more precise and efficient methods. Traditional diagnostic techniques often face challenges, such as human error, inconsistencies, and time constraints, that affect the accuracy and speed of diagnosis. Early detection of DR requires models that can capture complex patterns in retinal images. However, single models often struggle with the diversity and intricate nature of the images by leading to less reliable predictions. To address this, we propose an ensemble method that incorporates advanced image processing techniques and optimization strategies to improve model accuracy. Our methodology begins with a robust image pre-processing phase, which includes CLAHE to enhance the contrast of retinal images and Gamma correction to adjust brightness. Following this, we perform image fusion using DWT, which combines detail from images at multi-resolution to offer a more informative representation of the images by enabling better feature extraction. In the next phase, we select three high performing DL architectures including DenseNet169, MobileNetV1, and Xception, each contributing unique strengths in feature extraction. Further enhancing the feature extraction capability, we integrate an improved residual block after these models. This block addresses challenges like vanishing gradients and improves the capacity of models to capture more intricate patterns from retinal images. Finally, the predictions generated from these models are then aggregated using an intelligent ensemble approach, and we utilize the SSA to optimize the weight distribution among the base models. This method ensures that the contribution of each model is intelligently adjusted to maximize the overall performance of the ensemble model. The key contributions of this study include the following:
\begin{itemize}
\item We introduce an effective approach that begins with CLAHE for contrast enhancement and Gamma correction for brightness optimization, followed by image fusion using DWT. This fusion creates a richer, more informative dataset that improves the ability of the model to detect DR.
\item The proposed ensemble method integrates three powerful pre-trained models including DenseNet169, MobileNetV1, and Xception, each offering unique feature extraction capabilities to enhance DR detection. We further improve feature extraction by incorporating an improved residual block with the pre-trained models, which addresses issues like vanishing gradients and enhances the ability of the model to capture subtle patterns in DR images.
\item The innovation of this work is to optimize the ensemble model by dynamically allocating weights to the selected base architectures using SSA. Traditional methods rely on fixed weights and overlook the contribution of individual models. SSA optimizes the contribution of individual model regarding their performance, enhancing the integration, robustness, and classification accuracy of the ensemble.
\item We validate the proposed model on the multiclass Kaggle APTOS 2019 dataset and achieved an accuracy of 88.52\%, which highlights the effectiveness of our ensemble approach. To improve interpretability, Performance is further evaluated using various metrics, including the Precision-Recall curve, confusion matrix, and ROC curve plots, which collectively emphasize the model's accuracy and transparency.   
\end{itemize}
\section{Related work}\label{sec2}
DL has significantly advanced the diagnosis of DR by exhibiting high accuracy through complex neural networks. Integrating multiple DL models has further improved the performance by highlighting the significance of hybrid techniques. Several studies demonstrate strong potential in addressing the challenges of DR classification such as Mondal et al. \cite{mondal2022edldr} proposed an ensemble of modified DenseNet101 and ResNeXt models for DR classification. Their method was evaluated on APTOS19 dataset by using CLAHE for preprocessing with GAN-based augmentation to address class imbalance and achieved 86.08\% accuracy among five-classes. Wong et al.\cite{wong2023diabetic} proposed a Transfer Learning approach using ShuffleNet and ResNet-18 with an Error Correction Output Code (ECOC) ensemble for diabetic retinopathy detection. Adaptive Differential Evolution (ADE) was applied for feature selection and parameter tuning. Their method was evaluated on the APTOS dataset and achieved 82\% accuracy for five-class DR grading. 

Kobat et al. \cite{kobat2022automated} proposed a patch-based deep-feature engineering model for DR classification using DenseNet201 for feature extraction and neighborhood component analysis for feature selection, with a cubic SVM for classification. Their method was evaluated on the 5 class APTOS 2019 dataset and achieved 87.43\% accuracy. Oulhadj et al. \cite{oulhadj2024diabetic} proposed a hybrid DL  method combining fine-tuning vision transformer and a modified capsule network for DR severity level prediction. Their approach was evaluated on the APTOS dataset and achieved 88.18\% accuracy for severity level classification. Oulhadj et al. \cite{oulhadj2024enhancement} proposed an automatic method for DR detection using DenseNet-121 and a capsule network, along with discrete wavelet transform in the preprocessing step. Their method was evaluated on the APTOS dataset by achieving an 86.72\% accuracy. Macsik et al.\cite{macsik2024image} developed an ensemble deep learning model combining Xception and EfficientNetB4 architectures for diabetic retinopathy classification by utilizing multiple preprocessed image datasets. Their model achieved 88.36\% accuracy in classifying the five stages of DR by evaluating on the APTOS dataset. 

Bodapati et al. \cite{bodapati2024adaptive} developed an Adaptive Ensemble Classifier that utilizes deep spatial representations from multiple pre-trained CNNs for classifying the severity of DR. Their method was evaluated on the APTOS 2019 dataset, attaining 81.86\% accuracy by demonstrating superior performance compared to uni-modal representations. Yue et al. \cite{yue2023attention} introduced an Attention-Driven Cascaded Network (ADCNet) for DR classification, which employs a hybrid attention module to extract lesion-aware information without the need for manual annotation. Their model on APTOS dataset showed 83.40\% accuracy across multiclass. Mandiga et al.\cite{mandiga2022retinal} developed a DL architecture based on MobileNet for predicting diabetic retinopathy from retinal fundus images. Their model was tested on the APTOS 2019 dataset that showed 84\% accuracy in classifying diabetics. The methods discussed above show promising results in DR classification, but there is still room for further improvement. To enhance the robustness of these models, further optimization techniques need to be explored. By optimizing the contributions of each DL model, we can develop a more effective ensemble approach that improves the overall performance.

\section{Methods and materials}\label{sec3}
This section describes the methodology for DR classification using retinal fundus images. We begin by collecting a comprehensive dataset of DR images, which serves as the foundation of our approach. In the data preprocessing phase, we apply several techniques such as resizing, normalization, and data-augmentation to improve the diversity of datasets. Additionally, we incorporate image fusion using the DWT to capture features at multi-resolution that strengthen the capability of models to detect DR at different stages. Our methodology focuses on three pre-trained architectures, such as DenseNet169, MobileNetV1, and Xception. The features extracted from these base architectures are further strengthened by incorporating an improved residual block with each base architecture. These models are the core components of our ensemble strategy. Finally, we optimize the ensemble model using the Salp Swarm Optimization algorithm (SSOA), which combines the predictions from each model to improve classification accuracy. This combination of advanced feature extraction and optimization systems ensures an efficient method for DR classification.
\subsection{Dataset Description}\label{subsec2}
The APTOS 2019 dataset \cite{saproo2024deep} comprises 5,590 fundus images, with 3,662 of them made publicly available for training purposes. These images are labeled according to the international diabetic retinopathy (DR) grading standard, which categorizes DR into five distinct classes. For our experiments, we split the 3,662 accessible images into a train and a test set. The training set includes 90\% of the images from each class, while the test set consists of the remaining 10\%. This split ensures that each class is well-represented in both the training and testing sets, enabling robust model evaluation and minimizing class imbalance during training.

\subsection{Image fusion with Discrete Wavelet Transformation}\label{subsubsec2}
In the image fusion process depicted in Fig 1, the original image undergoes enhancement through CLAHE to improve contrast and detail visibility. Following this, Gamma correction is applied to adjust the brightness and optimize the image for better feature recognition. These enhanced images are then fused using the DWT, which combines multi-resolution details from both images. The DWT captures and integrates features from different scales by providing a richer, more informative image for subsequent analysis and improving the ability of models to detect subtle patterns in DR.
\begin{figure}[h]
    \centering
    \includegraphics[width = 8cm]{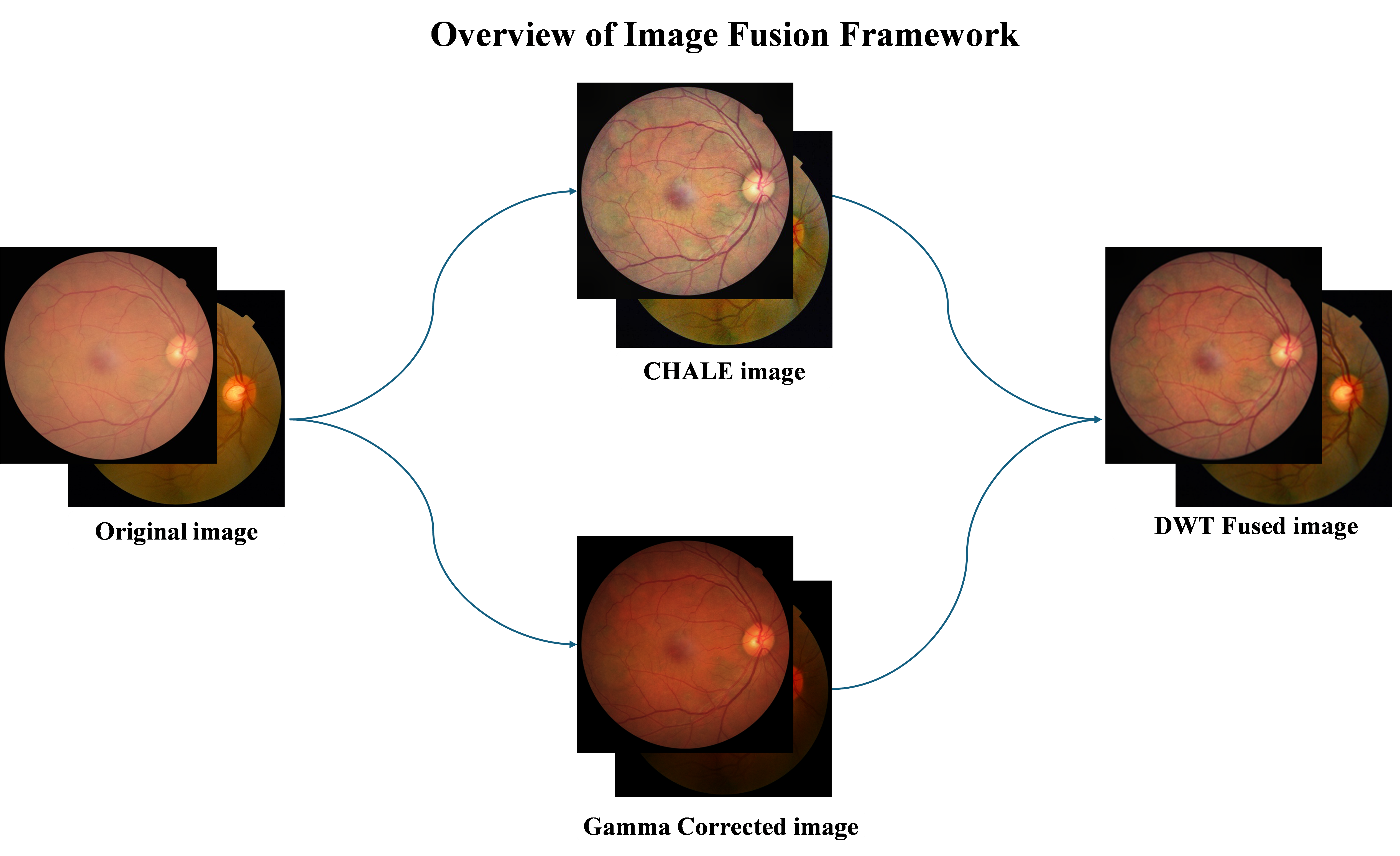}
    \caption{Overview of image Fusion Framework}
    \label{fig:se.png}
\end{figure}
\subsection{Selection of base models for feature extraction}\label{sec4}
The baseline selection for this study was based on the high performance observed during our initial experimental setup. We evaluated various pre-trained models, focusing on those that demonstrated superior accuracy and efficiency in early tests. The chosen baselines exhibit strong reliability and consistency in handling diabetic retinopathy detection. These selection ensures that our proposed method is compared against well-established and high-performing benchmarks.
\subsubsection{DenseNet169 architecture}
DenseNet169 \cite{khan2024hybrid} is a CNN architecture, which is designed to enhance the reuse and gradient flow of feature via dense connections between the layers. In this architecture, each layer receives input from all previous layers and then feeds its feature maps to all later layers within the dense block. This connection pattern boosts efficient feature extraction and lowers the number of parameters, as each layer only ought to generate a small set of feature maps. The number 169 indicates the total number of layers by making it a deep but computationally efficient model. By promoting feature diversity and alleviating the vanishing gradient problem, it excels at capturing complex patterns in the data while keeping the usage of memory relatively low. Its ability to achieve high performance with fewer parameters considers it a strong backbone architecture for tasks such as image classification, object detection, and segmentation, as well as a robust candidate for transfer- learning applications.
\subsubsection{MobileNet architecture}
MobileNetV1 \cite{howard2017mobilenets} is a lightweight CNN designed for mobile and embedded vision applications where computational resources are severely constrained. The key innovation of MobileNetV1 is the use of depthwise separable convolutions, which factorizes the standard convolution into two stages: depthwise separable convolutions that apply a filter to each input channel, followed by pointwise convolutions (1x1 convolutions) that combines the outputs across channels. This approach significantly decreases the number of parameters and computational cost compared to traditional convolutional operations, making the model faster and more efficient. Despite its lightweight design, it achieves highly competitive performance on image classification related tasks. As a result, it has become a popular choice. Its performance and speed make it an excellent benchmark for scenarios where balancing accuracy and computational cost is critical.
\subsubsection{Xception Architecture}
Xception \cite{chollet2017xception} is a state-of-the-art CNN architecture that further extends the design concept of the Inception network. The core innovation of Xception is to replace the standard Inception module with depthwise separable convolutions. This convolution method is more efficient and effective in capturing spatial and cross-channel correlations. Depthwise separable convolutions split the operation into two parts. Firstly, a channel-by-channel convolution (depthwise convolution), which applies filters to each input channel separately to capture spatial features. Secondly, a pointwise convolution (1x1 convolution), which combines the outputs of each channel. This design significantly lowers the number of parameters and computational complexity while maintaining performance. In addition, Xception also incorporates residual connections from ResNet to alleviate the gradient vanishing problem and facilitate the training of very deep networks. Xception achieves state-of-the-art performance in image classification tasks and typically requires fewer parameters than traditional CNNs, making it an ideal baseline model for tasks that require high accuracy with less computational resources. Its ability to strike a balance between depth, accuracy, and resource usage makes it a preferred choice in a variety of computer vision applications.
\subsection{Improved Residual block}
The original residual block \cite{he2016deep} is a key component in deep neural networks to address the vanishing gradient problem by introducing skip connections that allow the gradient to flow more easily through the network during backpropagation. This block typically uses the ReLU activation function to introduce non-linearity. However, we improved this residual block by replacing the ReLU activation function with the Leaky ReLU activation function. The Leaky ReLU allows a small, non-zero gradient when the unit is not active, which helps to prevent dead neurons and ensures better gradient flow. This modification enhances the ability of networks to learn from data and improves overall performance in deeper networks.Fig 2 presents the overview of original, and improved residual block.
\begin{figure}[h]
    \centering
    \includegraphics[width = 8cm]{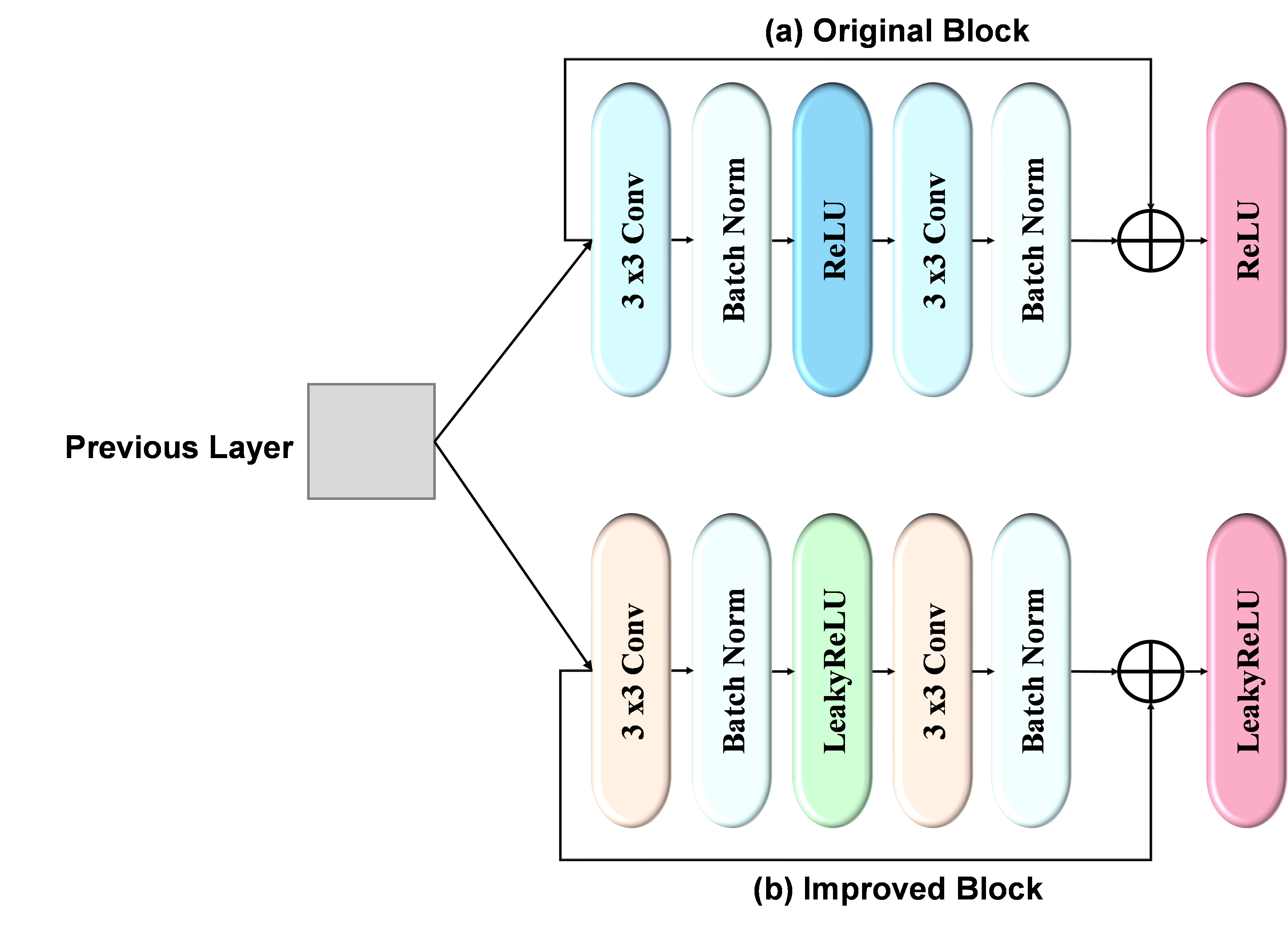}
    \caption{Architecture overview of original (a) \& improved block (b)}
    \label{fig:se.png}
\end{figure}
\subsection{Optimization with Salp Swarm Algorithm (SSA) strategy}
The SSA is a metaheuristic optimization technique inspired by the collective behavior of salps and proposed by Mirjalili et al.\cite{mirjalili2017salp} in 2017. In SSA, the population of salps is categorized into two distinct roles: leaders and followers. The leading salp directs the entire group toward the best possible solution, while the followers dynamically change their positions by following either the leader or other nearby followers. This process allows the algorithm to balance exploration and exploitation. Initially, it explores the search space globally and later refines the search in a more localized manner. The SSA is particularly effective due to its simplicity and efficiency, making it suitable for solving complex optimization problems. It has been proven to be effective in tackling a wide range of challenges, including optimizing functions, fine-tuning parameters, and solving engineering design problems. In this approach, every potential solution is modeled as a salp, and its movement is guided by three key components: the global optimal solution discovered, the local optimal solution within its neighborhood, and a random factor that introduces variability.

The SSA algorithm effectively handles noise and outliers by maintaining reliable performance in inconsistent or noisy environments. This robustness makes SSA particularly suitable for optimization problems in practical applications, where ensuring accuracy is crucial despite imperfections in the data. Although there is limited mathematical modeling of swarming behaviors in salp populations, SSA offers a unique approach. Unlike other swarm-based algorithms, such as those inspired by bees, ants, or fish, no specific model has been developed for salp swarms. In the SSA, the salps are distributed across an n-dimensional search space, and their locations are organized in a 2-dimensional matrix. Within this space, a food source is considered to be present, serving as the goal for entire swarm. The position of the leader salps is adjusted using a predefined mathematical formula, which helps to guide the swarm toward the target.

The following equation (1) updates the position of the leader salp based on the food source location. 
The position of the leader salp in the \(j^{\text{th}}\) dimension is given by:

\[
S_j^i = \begin{cases}
F_j + a\left((ub_j - lb_j) \cdot b + lb_j\right) & \text{if } c \geq 0.5 \\
F_j - a\left((ub_j - lb_j) \cdot b + lb_j\right) & \text{if } c < 0.5
\end{cases}
\tag{1}
\]

Where:
\[
S_j^i : \text{Position of the leader salp in the } j^{\text{th}} \text{ dimension},
\]
\[
F_j : \text{Position of the food source in the } j^{\text{th}} \text{ dimension},
\]
\[
ub_j : \text{Upper bound of the } j^{\text{th}} \text{ dimension},
\]
\[
lb_j : \text{Lower bound of the } j^{\text{th}} \text{ dimension},
\]
a, b, c: \text{Randomly generated values, where } a \text{ controls exploration vs. exploitation, and } b, c \text{ are uniformly distributed in } [0,1]. The coefficient \(a\) plays a vital role in SSA, as it controls the trade-off between exploration and exploitation. It is mathematically represented as:
\[
a = 2e^{\left(-\left(\frac{t}{T}\right)^2\right)}
\tag{2}
\]

Where:
\[
t : \text{Current iteration},
\]
\[
T : \text{Total number of permitted iterations},
\]
\[
a : \text{Decreases over time by balancing exploration and exploitation}.
\]

The follower salps change their positions relative to the positions of salps ahead of them, as shown in:

\[
P_j^i = \frac{1}{2} \left( P_j^i + P_j^{(i-1)} \right)
\tag{3}
\]

Where:
\[
P_j^i : \text{Updated position of the } i^{\text{th}} \text{ salp in the } j^{\text{th}} \text{ dimension},
\]
\[
P_j^{(i-1)} : \text{Position of the salp immediately ahead of the } i^{\text{th}} \text{ salp in the } j^{\text{th}} \text{ dimension}.
\]
The step-by-step process of the SSO method is visually represented in Fig 3 through a flowchart, which breaks down each stage of the algorithm to ensure clarity and enhance comprehension.
\begin{figure}[h]
    \centering
    \includegraphics[width = 12cm]{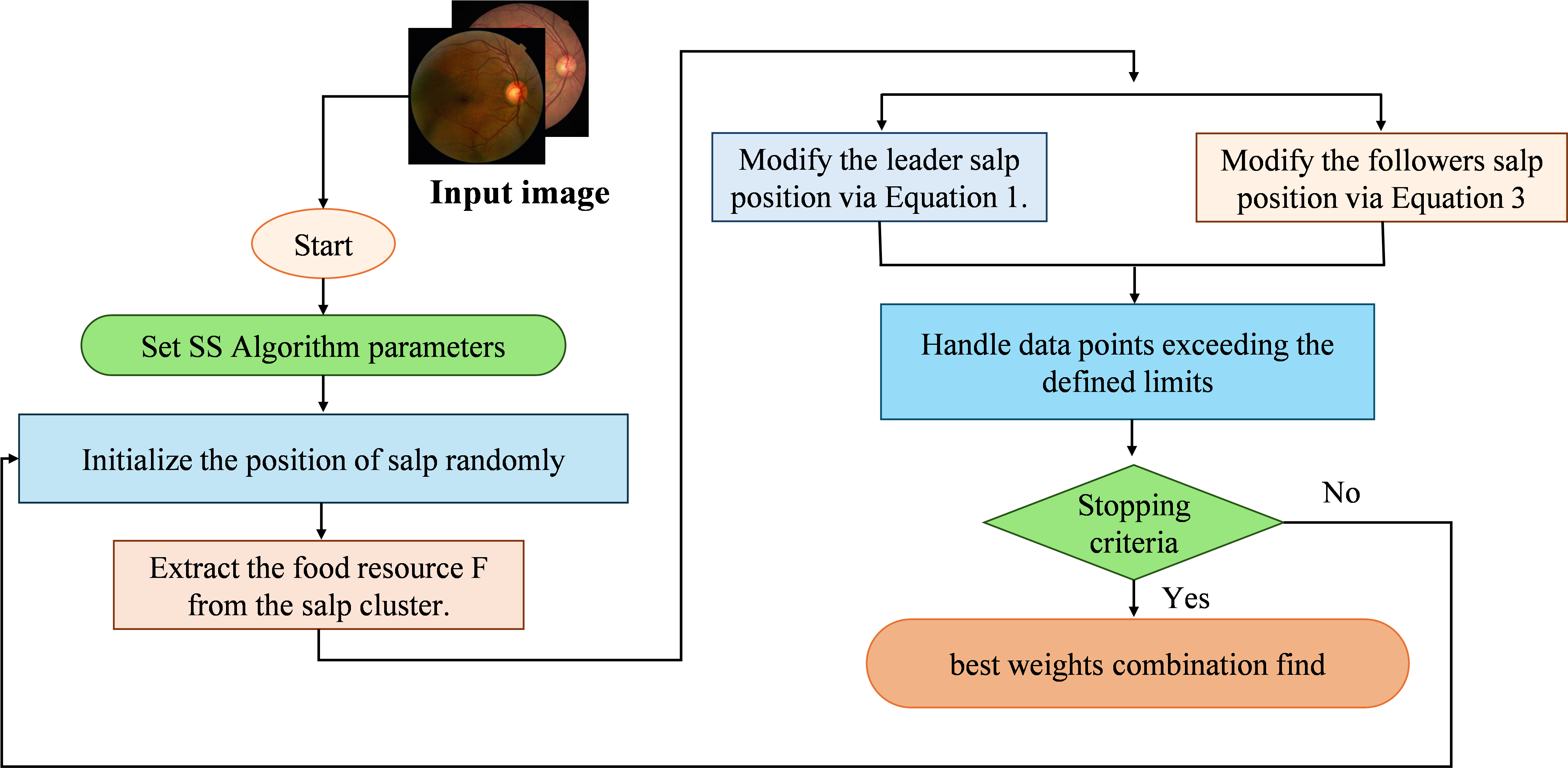}
    \caption{Flowchart illustrating the step-by-step process of SSA}
    \label{fig:se.png}
\end{figure}
\subsubsection{Objective Function for Ensemble Optimization}
This research explores the influence of different weight allocations on the performance of the ensemble model by establishing a critical objective function. The method utilizes the SSA to search for the best weights combination by combining predictions from three base architectures through a weighted averaging approach. The main objective is to enhance the accuracy of the models. The detailed process for evaluating the objective function is as follows:\\\\
\textbf{Step 1: Weighted Aggregation of Predictions}
First, we calculated the weighted outputs by taking the average of the predictions from each base model, using their respective assigned weights as the basis for the computation. The formulation for this computation can be expressed as in Equation (4):
\[
W_{\text{pred}} = \sum_{i=1}^{N} w_i \cdot \hat{p}_i \tag{4} \]
Here, $w_i$ represents the weight assigned to the $i$-th model. These weights dictate the contribution of each individual model to the ensemble. $\hat{p}_i$ represents the class probabilities predicted by the $i$-th model, where each value indicates the likelihood of a specific class as predicted by that model. $N$ represents the total number of base models in the ensemble. The $\mathbf{W}_{\text{pred}}$ is the combined weighted prediction vector from all models.\\\\
\textbf{Step 2: Aggregation of Weighted Predictions}
The following step involves aggregating the weighted outputs to establish the final class labels of the ensemble model. This process is carried out using Equation (5).
\begin{equation}
\text{Aggregated}_{\text{pred}} = \arg\max(W_{\text{pred}}, \, \text{axis} = 1) \tag{5}
\end{equation}
\textbf{Step 3: Retrieve Actual Class Labels}
Subsequently, the true class labels for the test dataset are acquired using Equation (6).
\[ Y_{\text{true}} = \text{test\_data.labels} \quad \text{(6)} \]
\textbf{Step 4: Compute Accuracy}
Finally, the overall accuracy of the model is calculated by comparing the predicted class labels to the actual labels:
\begin{equation}
\text{Accuracy} = \text{accuracy\_score}(Y_{\text{true}}, \text{Aggregated\_pred}) \tag{7}
\end{equation}
This metric evaluates the performance of the ensemble model by assessing its ability to accurately classify the test data by considering the optimized weight distribution assigned to each base model.
\subsubsection{Parameter-Tuned Optimization Process for Ensemble Weight Adjustment}
The SS algorithm is employed to optimize the distribution of weights for the ensemble model by searching through the weight space within the range of [0-1]. The objective is to determine the best weight combination that enhances accuracy. The optimization process relies on the parameters outlined in Table 1, with the aim of identifying the weight distribution that maximizes the classification accuracy of the ensemble model. 
\begin{table}[h]
\centering
\caption{Parameters for optimization}
\begin{tabular}{cc}
\hline
\textbf{Parameters} & \textbf{Values} \\
\hline
Number of Salps (NS) & 100 \\
Max\_no\_iterations (Maxitr) & 100 \\
Number of weights to optimize (WN) & 3 \\
Upper limit of weight (Ul) & 1 \\
Lower limit of weight (Ll) & 0 \\
\hline
\end{tabular}
\end{table}
\subsection{Overview of Proposed Methodology Framework}
The proposed method for DR detection follows a multi-step approach designed to optimize the accuracy and robustness of the proposed model. It initiates with the pre-processing of diabetic input images by applying CLAHE and gamma correction . CLAHE enhances the contrast of the images by limiting the over-amplification of noise in homogeneous regions of the diabetic disease image, while gamma correction ensures that the pixel values are adjusted to a perceptually uniform brightness, thus enhancing the visibility of crucial features in the images. These preprocessing techniques are essential because they allow DL models to better capture and learn the significant patterns related to DR. Following image enhancement, Discrete Wavelet Transformation technique is applied to fuse the enhanced images. DWT is used because it is highly effective in capturing both high and low-frequency components of the images, which are significant for extracting detailed features from medical images. The fusion process enables the model to learn a more comprehensive representation of the image data by improving the overall feature extraction process. Once the images are preprocessed and enhanced, the dataset is split into training and testing subsets. This ensures that the model is trained on a portion of the data and validated on an unseen subset, which is crucial for assessing its generalization capability. We then trained several pre-trained DL models on specific diabetic images dataset to identify the top performing model. These pre-trained models have shown strong performance in various image classification tasks and have been proven effective in extracting complex features from medical images. By leveraging these models, we take advantage of their ability to capture relevant features from large-scale datasets. After the training phase, we evaluate the performance of each pre-trained model on the DR images dataset and select the three top performing models based on their accuracy and other relevant metrics. This ensures that only the best-performing models contribute to the final prediction by reducing the risk of underperforming models impacting the results. After performance comparison we selected DenseNet169, MobileNetV1, and Xception architectures. These architectures are preferred due to their exceptional performance on our specific diabetic images dataset. To further enhance the feature extraction capability of these architectures, we leveraged an improved residual block. Residual blocks are commonly used to mitigate the vanishing gradient problem, which is especially important in deep networks. The improved residual block helps to improve the flow of gradients through the network by enabling the model to learn more effectively even in deeper layers. This modification significantly boosts the performance of the selected models by ensuring better feature extraction from the DR dataset.
Once the models are trained and their predictions are generated, these predictions are passed to the Salp Swarm Algorithm (SSA). SSA is employed to optimize the weight assignments of each model in the ensemble by allowing the proposed model to adaptively learn the contribution of each model based on its individual performance. This step is critical because it ensures that the contribution of each model is maximized by leading to improved overall accuracy. The SSA algorithm works by simulating the foraging behavior of salps, which helps in finding optimal solutions in a highly dynamic environment like ensemble learning.

Finally, to assess the robustness and reliability of the proposed method, we evaluate its performance using several key metrics including accuracy,precision,recall,F1-score, confusion matrix, ROC curve, P-R curve, and McNemar's statistical test. The accuracy metric provides an overall performance measure of the model, while the confusion matrix helps to identify misclassifications between the different classes. The ROC curve and PR curve allow for a deeper understanding of the trade-offs between true positive and false positive rates, as well as the precision and recall of the model. McNemar’s test is conducted to evaluate the statistical significance of the differences between the proposed model and the base models pairwise by ensuring that the improvements observed are not due to random chance. This comprehensive approach ensures that the proposed method is robust, accurate, and well-suited for the detection of DR in medical images. The overall block diagram of Proposed Methodology with crucial components is illustrated in Fig 4.
\begin{figure}[h]
    \centering
    \includegraphics[width = 12cm]{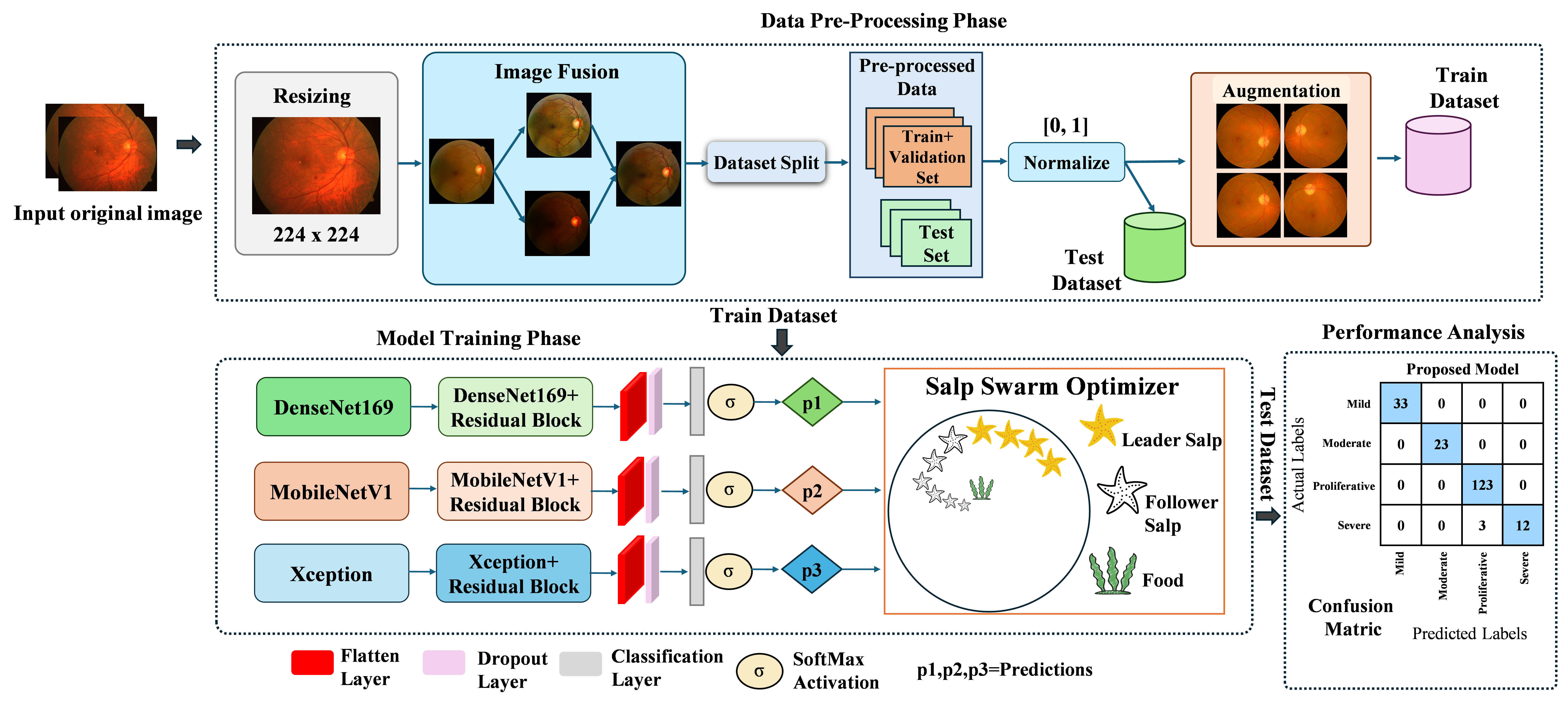}
    \caption{Illustration of the Proposed Methodology for Diabetic Retinopathy Detection Using the SSA-Driven Ensemble approach}
    \label{fig:se.png}
\end{figure}
\section{Implementation and Results}\label{sec5}
This section outlines the experimental setup for diabetic retinopathy (DR) detection, including evaluation metrics, preprocessing steps, augmentation techniques, and model training hyperparameters. We measure the performance of proposed model by comparing it with different pre-trained CNN architectures and existing study. An ablation study is also conducted to highlight the contribution of each technique, demonstrating the effectiveness of our approach in DR classification.
\subsection{Experimental setup and Hyperparameter details}
The model developed in this study was implemented using Python and the Keras framework. All experiments were carried out in a Jupyter notebook within the Anaconda environment, utilizing GPU support for enhanced computational efficiency. The model training and evaluation were executed on an NVIDIA GeForce MX350 Tesla GPU with 8 GB of RAM. Hyperparameters are crucial for fine-tuning the performance of models, as they control various aspects of the training process, such as learning rate, batch size, and activation functions, which directly influence the ability of models to converge and generalize effectively. Table 2 outlines the key training parameters for the model. The input images were resized to a consistent 224 × 224 pixels. We employed the optimizer named Adam with a learning rate of 0.001, which is preferred for its ability to adapt the learning rate dynamically and promote faster convergence. A batch size of 64 was chosen to strike a balance between computational efficiency and the effectiveness of the training process. Training was carried out for 20 epochs, with early stopping implemented to prevent overfitting once performance stabilized. The best-performing models were saved using the Model Checkpoint function. To address the vanishing gradient issue, LeakyReLU activation was used, while SoftMax was applied in the final layer to output class probabilities. For the multi-class classification task, categorical cross-entropy loss was employed as the loss function, while a dropout rate of 0.2 was used to mitigate the potential for overfitting.
\begin{table}[h]
\centering
\caption{Hyper-parameters}
\begin{tabular}{cc}
\hline
\textbf{Hyper Parameters} & \textbf{Value} \\ \hline
Input image Size          & 224×224        \\ 
Learning Rate            & 0.001          \\ 
Batch Size               & 64             \\ 
Dropout rate             & 0.2            \\ 
Learning Optimizer       & Adam           \\ 
Activation Function      & SoftMax        \\ 
Epochs                   & 20             \\ \hline
\end{tabular}
\end{table}
\subsection{Evaluation metrics}
In this study, the performance of the proposed model is evaluated using statistical metrics presented in the following Equations (8-11). These metrics are calculated based on the values obtained from the confusion matrix, which includes \( P_{\text{true}} \), \( N_{\text{true}} \), \( P_{\text{false}} \), and \( N_{\text{false}} \). The evaluation formulas are shown in the following equations (1-4).
\begin{equation}
\text{Accuracy} (A) = \frac{P_{\text{true}} + N_{\text{true}}}{P_{\text{true}} + N_{\text{true}} + P_{\text{false}} + N_{\text{false}}} \tag{8}
\end{equation}

\begin{equation}
\text{Precision} (P) = \frac{P_{\text{true}}}{P_{\text{true}} + P_{\text{false}}} \tag{9}
\end{equation}

\begin{equation}
\text{Recall} (R) = \frac{P_{\text{true}}}{P_{\text{true}} + N_{\text{false}}} \tag{10}
\end{equation}

\begin{equation}
\text{F1-Score} (F1) = \frac{2 \times (P \times R)}{P + R} \tag{11}
\end{equation}
\subsection{Pre-processing and augmentation}
The Data Pre-Processing and Augmentation Phase of the DR image analysis pipeline starts with the original image, which undergoes preprocessing steps like resizing, normalization, and contrast adjustment. To enhance the diversity of the training dataset, several augmentation techniques such as rotation, flipping, and cropping are applied. These augmented images are subsequently used to improve the generalization ability of models and to handle variations in the data by ultimately enhancing the performance in detecting DR, as shown in Fig 5.
\begin{figure*}[h]
    \centering
    \includegraphics[width = 10cm]{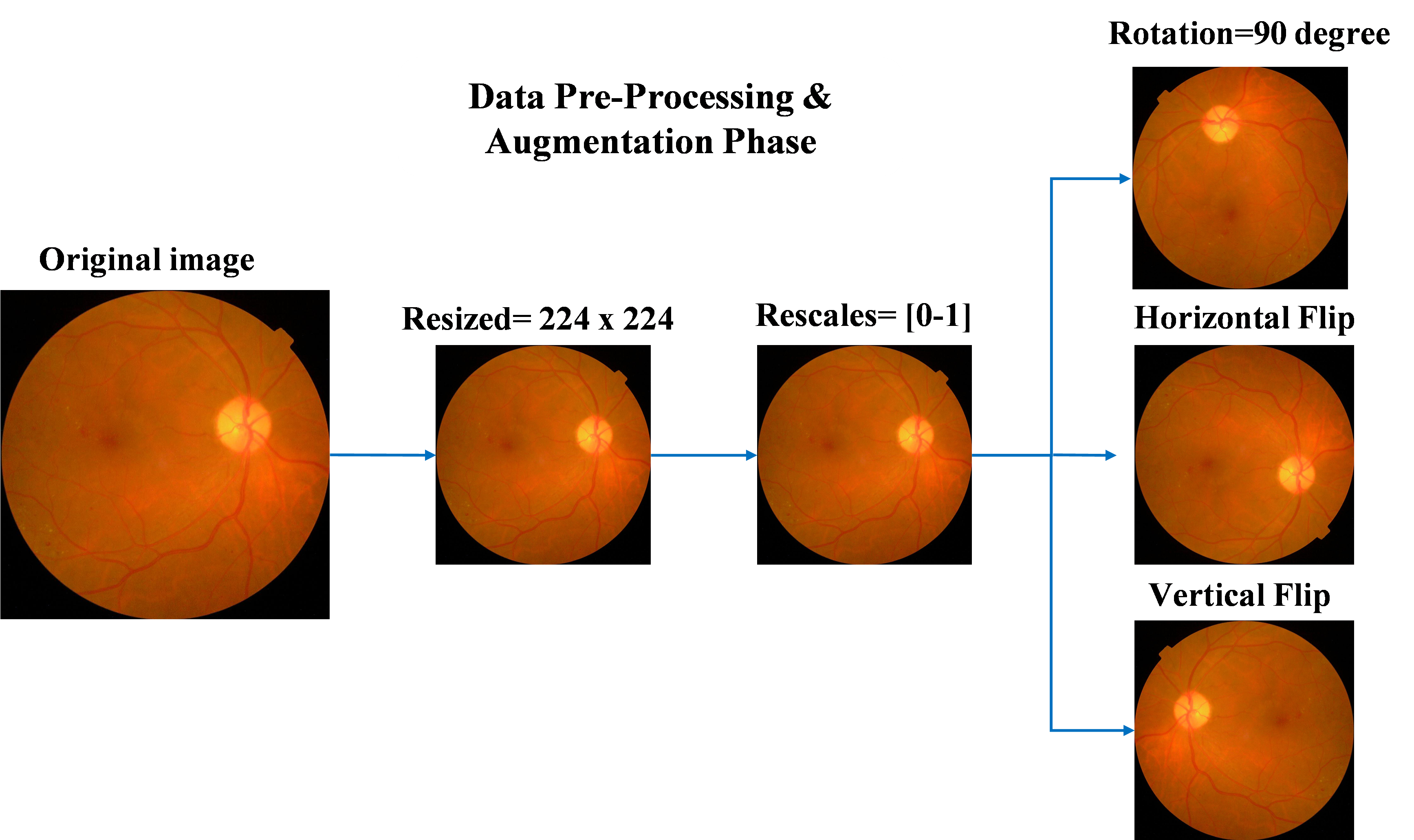}
    \caption{Illustration of the data pre-processing and augmentation steps for model training}
    \label{fig:se.png}
\end{figure*}
\subsection{Class wise performance analysis of Proposed model and enhanced baseline models}
The performance analysis of the proposed model and selected base models among each separate class on Kaggle APTOS 2019 dataset is presented in Table 3. The table compares the performance of each model across different classes of DR. It also demonstrates how the proposed model enhances classification accuracy for each class compared to the baseline models. For DenseNet169-Base model, the Moderate class shows the lowest precision at a value of 68.47\%, while the Normal class achieves the highest precision at 97.98\%. The precision values for the other classes range between 73.33\% to 80.00 \%. For recall, the Severe class also has the lowest value at 63.33\%, and the Normal class has the highest recall at 97.49 \%. The recall values for the other classes lie between 79. 41\%-87.36 \%. The overall accuracy achieved by DenseNet169-Base model is 86.06\%. For MobileNet-Base model, the Proliferative class shows the lowest precision value at 74.78 \% and Normal class shows the highest precision rate of 97.49 \%. Other classes exhibited precision values from 77.14 \%-79.79 \%. In terms of recall rate, the mild class showed the lowest value of 73.33 \% and Normal class exhibited 97.49 \% value. Other classes show values from 76.36 \% to 87.01 \%. The overall accuracy achieved by MobileNet-Base model is 85.51\% which is slightly lower than the first base model. For the Xception-Base model, the Moderate class shows the lowest precision value at 69.47 \% while the Normal class achieves the highest precision at 97.00\%. The precision values for the other classes range from 70.00\%, to 74.00\%. In terms of recall, the Mild class has the lowest value at 73.33\%, while the Normal class exhibits the highest recall at 97.49\%. The recall values for the other classes range from 75.29 \% to 78.48\%. The overall accuracy achieved by the Xception-Base model is 85.24\%, which is slightly lower than the first two base architectures.

For the Proposed Ensemble Model, the Moderate class shows the lowest precision value at 69.03\%, while the Normal class indicates the highest precision at 97.98\%. The Severe and Proliferative classes have the same precision value of 80.00\%. The mild class exhibited 84.00 \% precision value. In terms of recall, the Mild class has the lowest value at 83.33 \%, while the Normal class exhibits the highest recall at 97.49\%. The recall values for the other classes range from 85.29 \% to 89.66 \%. The overall accuracy achieved by the Proposed Ensemble Model is 89.07\%, which outperforms the individual base architectures.
\begin{table}[h]
\centering
\caption{Performance Metrics for Different Models}
\label{tab:performance}
\begin{tabular}{cccccc}
\hline
\textbf{Models} & \textbf{Class} & \textbf{Precision (\%)} & \textbf{Recall (\%)} & \textbf{F1-Score (\%)} & \textbf{Accuracy} \\ \hline
\multirow{5}{*}{DenseNet169 (Base 1)} & Mild & 80.00 & 79.50 & 74.55 & 86.06\% \\  
 & Moderate & 68.47 & 87.36 & 76.77 &  \\  
 & Normal & 97.98 & 97.49 & 97.73 &  \\ 
 & Severe & 73.33 & 63.33 & 65.83 &  \\  
 & Proliferative & 79.41 & 79.41 & 69.41 &  \\ 
\multirow{5}{*}{MobileNet (Base 2)} & Mild & 78.15 & 73.33 & 75.61 & 85.51\% \\ 
 & Moderate & 79.79 & 87.01 & 73.22 &  \\ 
 & Normal & 97.49 & 97.49 & 97.49 &  \\ 
 & Severe & 77.14 & 76.36 & 74.44 &  \\ 
 & Proliferative & 74.78 & 77.06 & 70.00 &  \\ 
\multirow{5}{*}{Xception (Base 3)} & Mild & 74.00 & 73.33 & 78.18 & 85.24\% \\ 
 & Moderate & 69.47 & 75.86 & 72.53 &  \\ 
 & Normal & 97.00 & 97.49 & 97.24 &  \\ 
 & Severe & 70.00 & 78.48 & 79.23 &  \\ 
 & Proliferative & 72.86 & 75.29 & 78.71 &  \\ 
\multirow{5}{*}{Proposed Ensemble Model} & Mild & 84.00 & 83.33 & 88.18 & 89.07\% \\
 & Moderate & 69.03 & 89.66 & 78.00 &  \\ 
 & Normal & 97.98 & 97.49 & 97.73 &  \\ 
 & Severe & 80.00 & 87.27 & 87.50 &  \\ 
 & Proliferative & 80.00 & 85.29 & 77.50 &  \\ \hline
\end{tabular}
\end{table}
A CM serves as a useful visual aid for evaluating classification performance, as it clearly displays the count of accurate and inaccurate predictions. Fig 6 provides a comprehensive comparison of the CM for the base and proposed ensemble model by emphasizing their respective classification performance. When comparing the base models, DenseNet169 (Base 1) in Fig 6(a) correctly classified 315 out of 366 test images by leaving 51 misclassifications. This indicates relatively high accuracy, with leaving room for improvement in terms of misclassification reduction. MobileNetV1(Base 2) in Fig 6(b) correctly classified 313 out of 366 images by resulting in 53 misclassifications, which are slightly more than DenseNet169 model. Xception (Base 3) in Fig 6(c) performed similarly by correctly classifying 312 out of 366 test samples, which led to 54 misclassifications. In contrast, the proposed ensemble model in Fig 6(d) demonstrated a significant improvement by correctly classifying 326 out of 366 test images by leaving only 40 misclassifications. This reduction in misclassifications across all classes highlights the strength of the ensemble model. By combining the strengths of each base model through advanced techniques like Salp Swarm Optimization (SSO) and image fusion, the ensemble approach mitigates the weaknesses of the individual models and effectively minimizes misclassifications with improved overall classification accuracy.
\begin{figure}[h]
\centering
\begin{minipage}[]{6cm}
  \centering
  \includegraphics[width = 5cm]{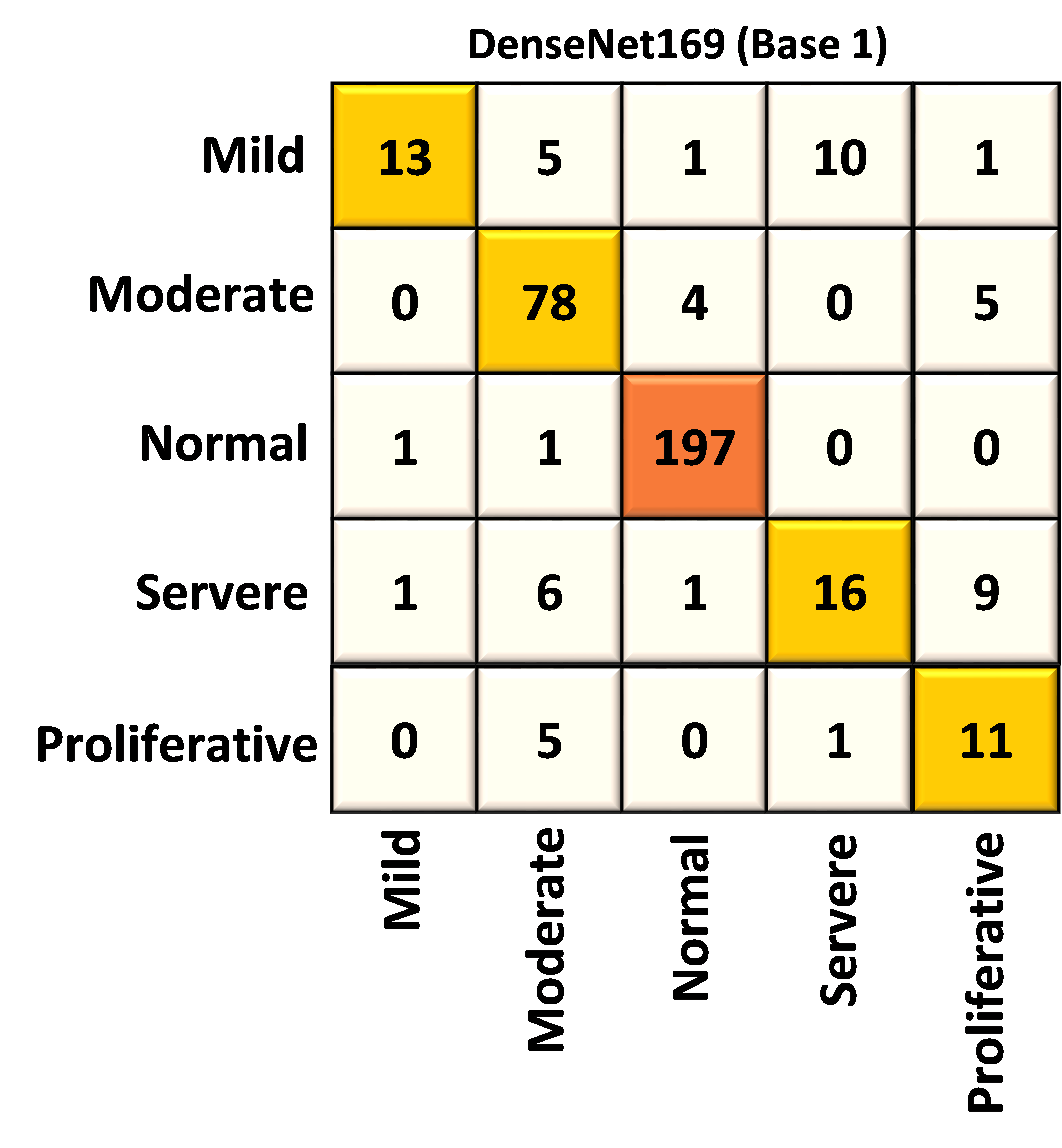}
  \vspace{-0.5cm} 
    \begin{center}
    \textbf{(a)}    
    \end{center}
\end{minipage}
\begin{minipage}[]{6cm}
  \centering
  \includegraphics[width = 5cm]{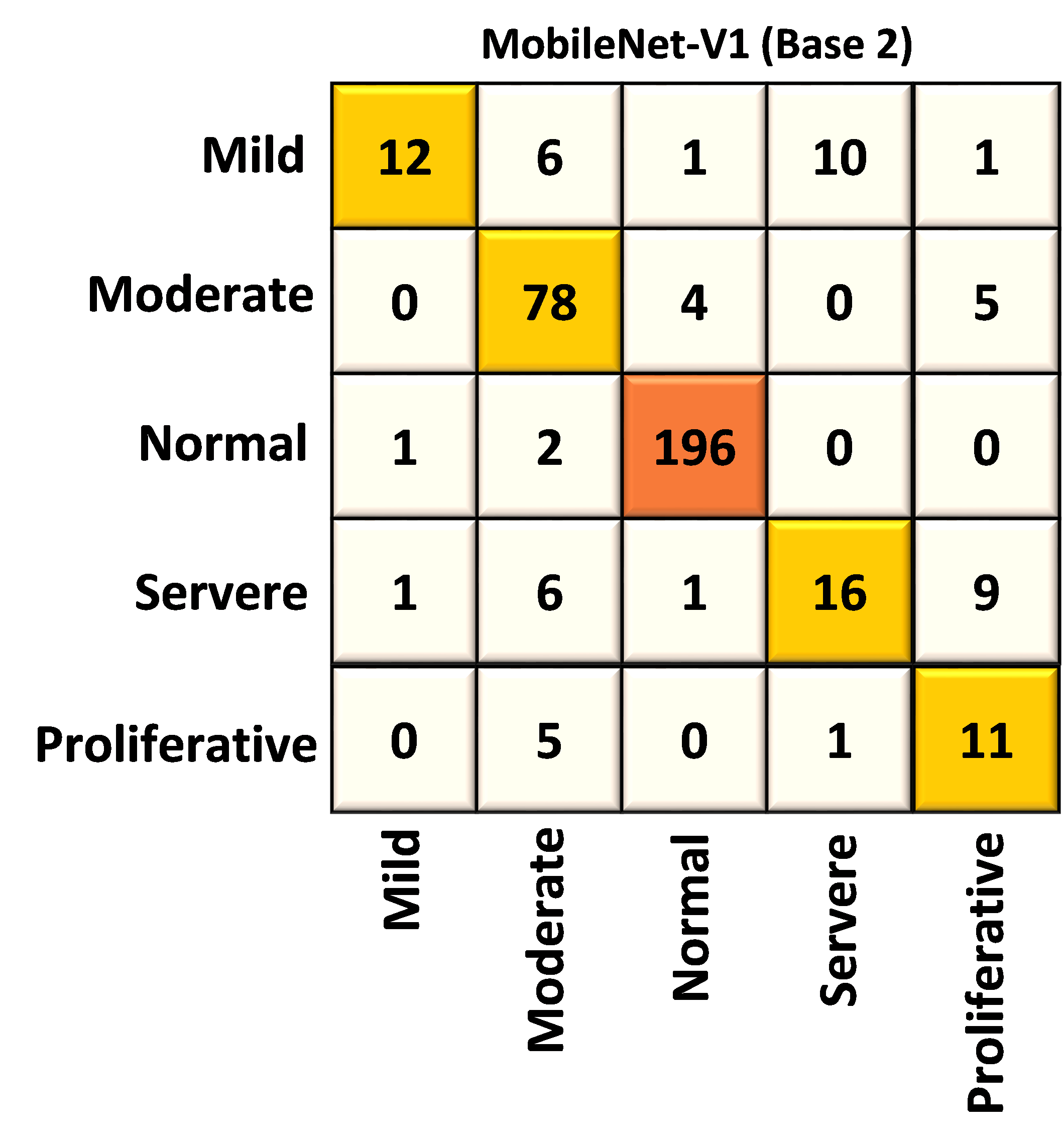}
  \vspace{-0.5cm} 
    \begin{center}
    \textbf{(b)}    
    \end{center}
\end{minipage}
  \begin{minipage}{6cm}
    \centering
    \includegraphics[width = 5cm]{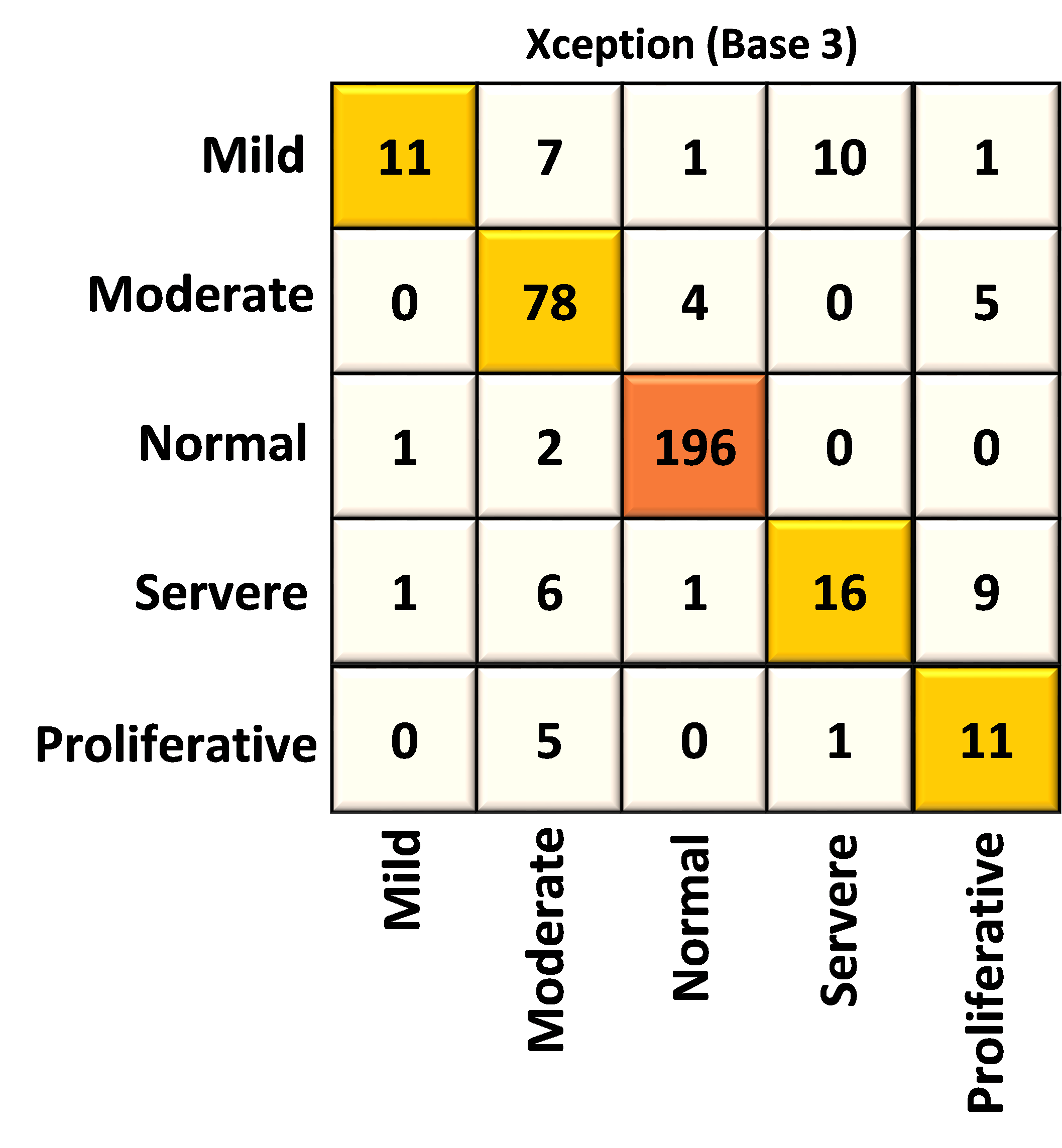}
    \vspace{-0.5cm} 
    \begin{center}
    \textbf{(c)}    
    \end{center}
  \end{minipage}
  \begin{minipage}{6cm}
    \centering
    \includegraphics[width = 5cm]{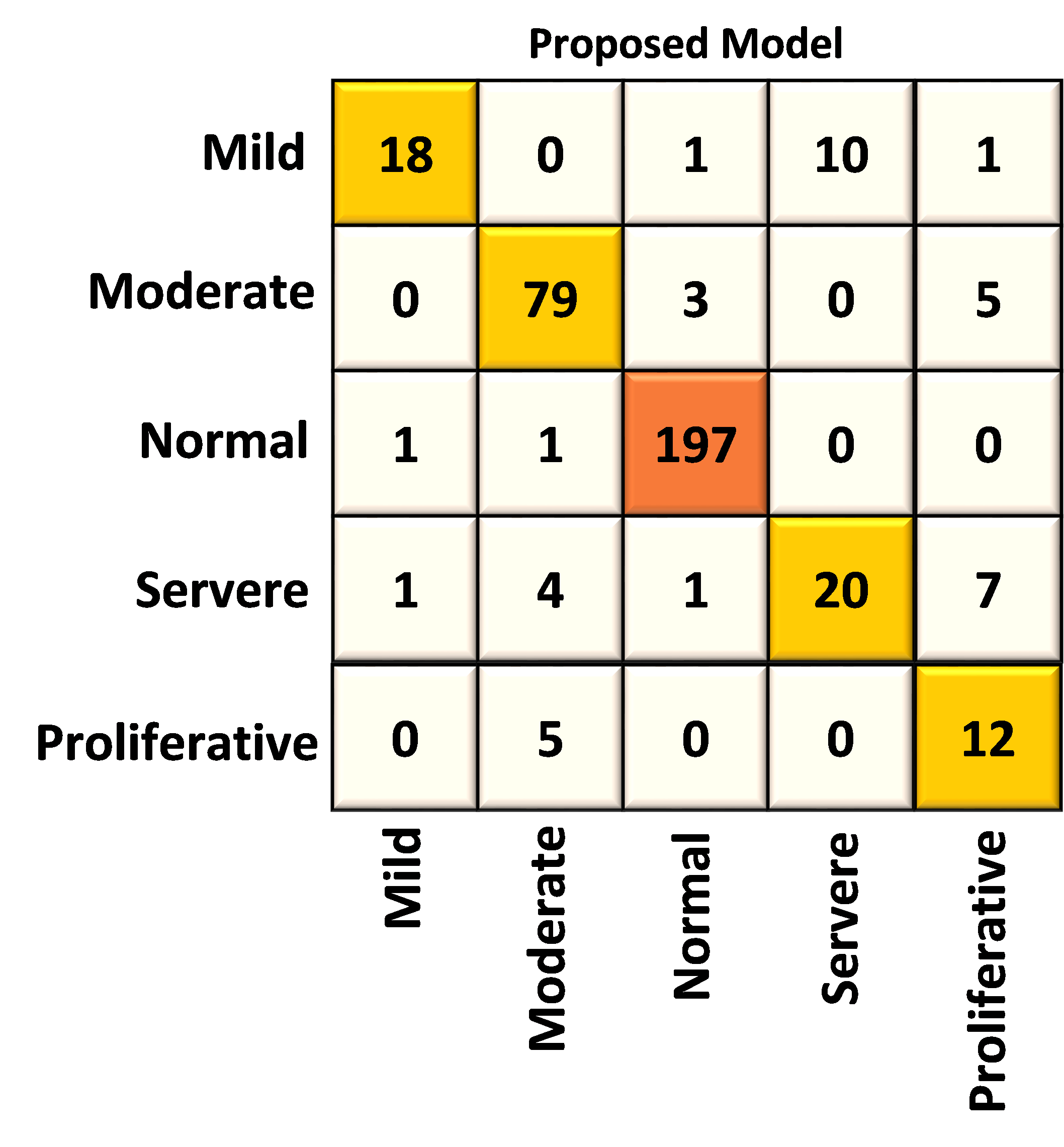}
    \vspace{-0.5cm} 
    \begin{center}
    \textbf{(d)}    
    \end{center}
  \end{minipage}
  \caption{Performance evaluation using CM for (a) DenseNet169 (Base 1), (b) MobileNetV1 (Base 2), (c) Xception (Base 3), and (d) the proposed ensemble model.}
\end{figure}
The ROC curve is a graphical tool used to assess the effectiveness of classification models. It illustrates the relationship between the True Positive Rate (TPR) and the False Positive Rate (FPR), demonstrating the balance between sensitivity and specificity at various thresholds. The area under the curve (AUC) offers a single metric that reflects the ability of models to differentiate between classes, where higher AUC values signify superior performance. Fig 7(a) compares the AUC values among four models. DenseNet169-Base achieves an AUC of 0.9614, followed by MobileNetV1-Base with an AUC of 0.9602, and Xception-Base with an AUC of 0.9636. The Proposed Model outperforms the single base architectures with an AUC of 0.9673 by indicating superior classification performance. This comparison reveals the enhanced performance of the proposed ensemble model to achieve higher true positive rates with minimal false positives.

The PR curve is a visual tool used to evaluate the effectiveness of classification models, emphasizing the trade-off between precision and recall at various thresholds. The Average Precision (AP) score is computed to offer a single metric that encapsulates the overall performance represented by the curve, with higher values indicating better performance. Fig 7(b) compares the Precision-Recall curves for the four models. DenseNet169-Base has an AP score of 0.8855, while MobileNetV1-Base achieves a slightly higher AP of 0.8953. Xception-Base further improves with an AP score of 0.8990. However, the Proposed Model outperforms all the base models with the highest AP score of 0.9060 by demonstrating superior precision and recall performance. This comparison illustrates the effectiveness of the proposed ensemble model in maintaining a higher level of precision while ensuring strong recall across varying thresholds, surpassing the performance of individual models.
\begin{figure}[h]
\centering
\begin{minipage}[]{6cm}
  \centering
  \includegraphics[width = 6cm]{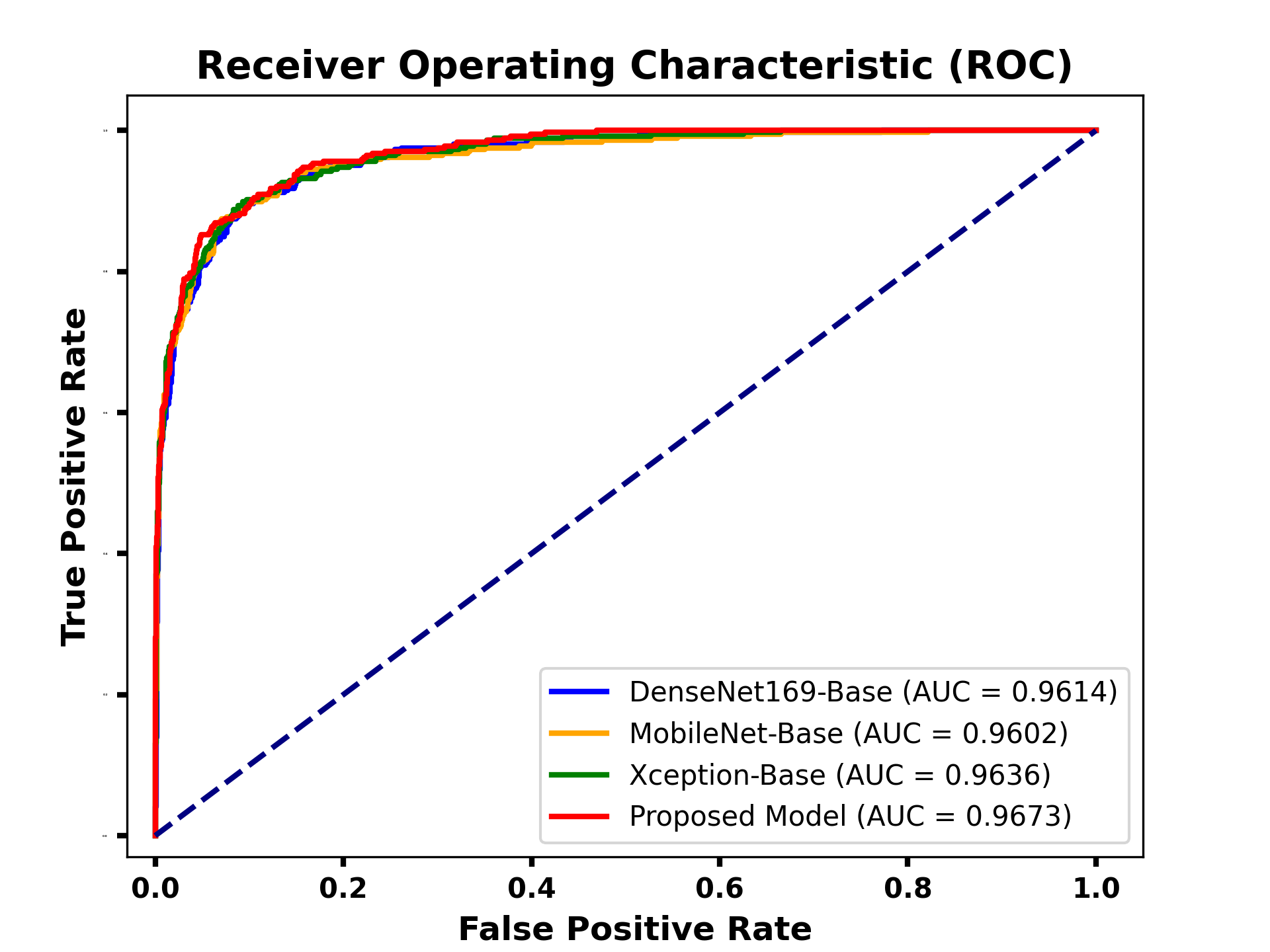}
  \vspace{-0.5cm} 
    \begin{center}
    \textbf{(a)}    
    \end{center}
\end{minipage}
\begin{minipage}[]{6cm}
  \centering
  \includegraphics[width = 6cm]{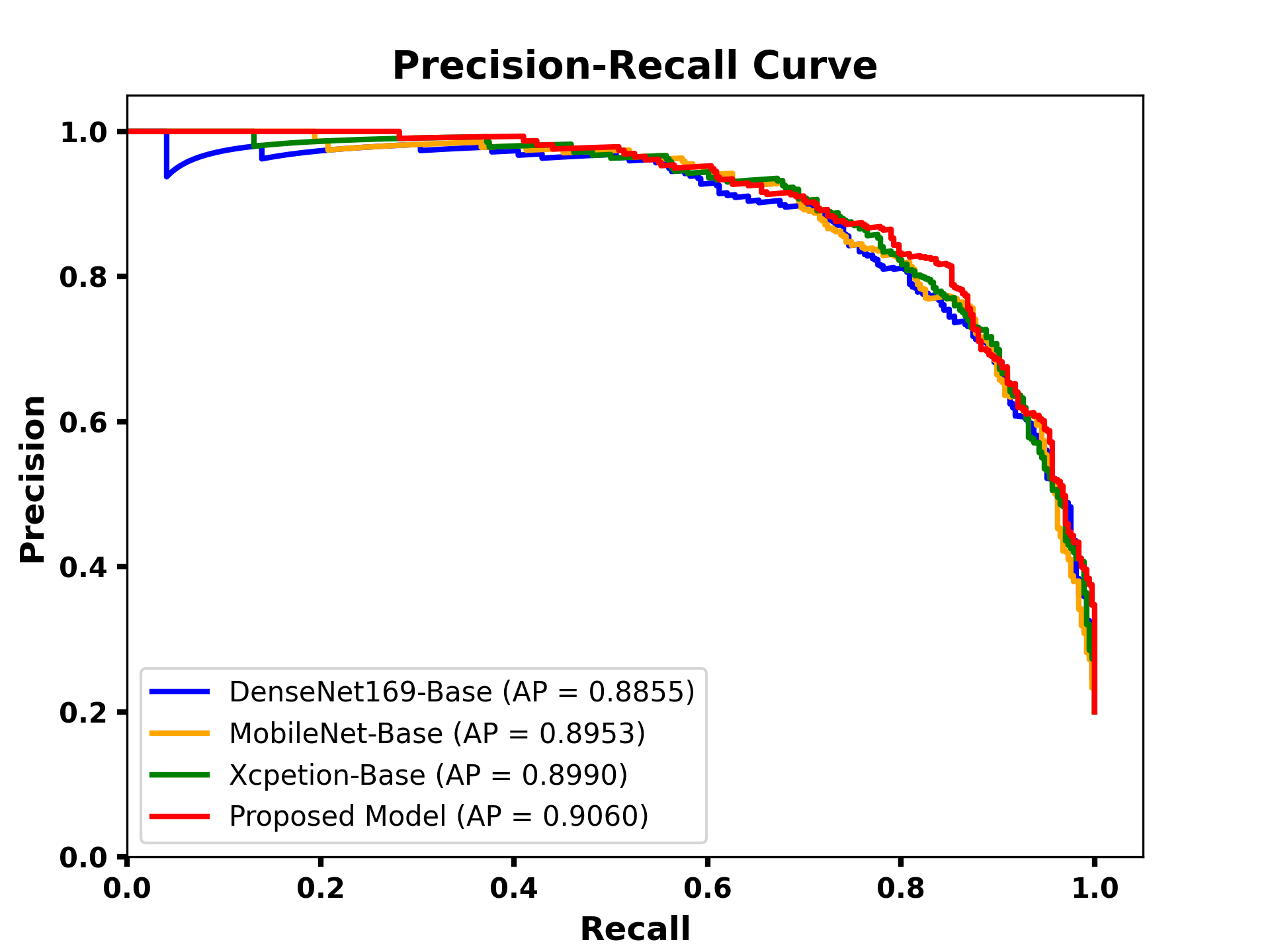}
  \vspace{-0.5cm} 
    \begin{center}
    \textbf{(b)}    
    \end{center}
\end{minipage}
  \caption{Performance comparison between the base models and the proposed ensemble model using (a) the ROC curve and (b) the Precision-Recall curve}
\end{figure}
\subsection{Performance analysis of Proposed model and existing CNN models}
We evaluate several baseline models using key performance metrics on the Kaggle APTOS 2019 dataset, as detailed in Table 4. The performance analysis of models reveals several interesting insights, particularly in terms of accuracy and recall. Among the baseline models, DenseNet169-Base stands out with the highest accuracy of 83.87\% and a strong recall score of 79.44\%, which indicates its effectiveness in both identifying DR cases and minimizing false positives. The MobileNetV1-Base model follows closely by achieving an accuracy of 82.78\% and a recall of 72.17\%, showing competitive performance but with slightly lower recall compared to DenseNet169. The Xception-Base model performs well with an accuracy of 83.06\% with lower recall of 71.37\%, particularly excelling in accuracy, although it slightly lags behind DenseNet169 and MobileNetV1 in recall score. Models like DenseNet121-Base and DenseNet201-Base exhibit strong performance with a similar accuracy of 80.05\%, which is lower than the top three selected base models but offers higher recall than both MobileNetV1 and Xception. The VGG family models show strong recall scores, but their accuracy falls behind the top-performing models. Meanwhile, the InceptionV3 and ResNet families present similar performance by exhibiting accuracy values around 77\%, with only slight variations. InceptionResNetV2-Base achieves the lowest accuracy among all models at 75.68\%, although its recall rate is the highest among all base models. Overall, DenseNet169-Base stands out as the most well-performed model by achieving both high accuracy and recall. The MobileNetV1 and Xception models also perform well in terms of accuracy, despite having slightly lower recall scores. These insights emphasize the need for balancing both accuracy and recall when selecting the optimal model for DR detection. While the proposed model has achieved highest performance as compared to listed CNN models.  
\begin{table}[h]
\centering
\caption{Performance comparison of proposed model with pre-trained models on Kaggle APTOS 2019}
\begin{tabular}{lcccc}
\hline
\textbf{Models} & \textbf{Precision (\%)} & \textbf{Recall (\%)} & \textbf{F1-Score (\%)} & \textbf{Accuracy (\%)} \\
\hline
DenseNet121-Base & 74.14 & 78.88 & 79.55 & 80.05 \\
DenseNet169-Base & 79.72 & 79.44 & 79.52 & 83.87 \\
DenseNet201-Base & 78.86 & 73.03 & 74.04 & 80.05 \\
InceptionV3-Base & 74.44 & 79.19 & 79.99 & 77.04 \\
InceptionResNetV2-Base & 72.05 & 79.91 & 70.16 & 75.68 \\
MobileNetV1-Base & 78.11 & 72.17 & 73.02 & 82.78 \\
MobileNetV2-Base & 77.97 & 75.75 & 75.62 & 79.50 \\
ResNet50V2-Base & 79.32 & 79.28 & 78.87 & 77.32 \\
ResNet101V2-Base & 74.49 & 71.28 & 71.46 & 77.04 \\
ResNet152V2-Base & 78.31 & 71.65 & 72.74 & 77.59 \\
VGG16-Base & 74.35 & 77.73 & 79.22 & 79.23 \\
VGG19-Base & 70.84 & 75.83 & 76.41 & 78.68 \\
Xception-Base & 74.13 & 71.37 & 71.71 & 83.06 \\
Proposed Model & 85.45 & 84.33 & 88.12 & 89.07 \\
\hline
\end{tabular}
\label{tab:comparison}
\end{table}
\subsection{McNamar’s statistical test analysis}
In order to statistically analyze the performance of our proposed ensemble method in comparison to the individual base models, we conducted McNemar's test. Table \ref{table:mcnemar} presents the p-values for each base model: DenseNet169 with a p-value of $1.47 \times 10^{-3}$, MobileNetV1 with a p-value of $2.60 \times 10^{-2}$, and Xception with a p-value of 0.0342. In McNemar’s test, a p-value below 0.05 indicates rejection of the null hypothesis. The p-values obtained are all below the significance threshold of 0.05, leading to the rejection of the null hypothesis for all models. This signifies a statistically significant difference in performance between the proposed ensemble method and the individual base models. The low p-values for DenseNet169 and Xception demonstrate highly significant differences, while the slightly higher p-value for MobileNetV1 still suggests a significant performance improvement. These results confirm the effectiveness of our proposed ensemble method in leveraging the complementary strengths of base models, resulting in superior predictive performance.
\begin{table}[h]
\centering
\caption{McNemar’s test}
\begin{tabular}{lc}
\hline
\textbf{Models} & \textbf{p-value} \\ \hline
DenseNet169-Base & $1.47 \times 10^{-3}$ \\ 
MobileNetV1-Base & $2.60 \times 10^{-2}$ \\ 
Xception-Base & 0.0342 \\ \hline
\end{tabular}
\end{table}
\subsection{Performance analysis of Proposed model with existing studies}
The performance comparison of the proposed optimized ensemble model with several existing methods on the APTOS 2019 dataset is outlined in Table 6. The comparison analysis of the existing methods reveals that the proposed model outperforms all other approaches on the APTOS 2019 dataset. While Xception method showed an accuracy of 86.40\%, the proposed method surpassed this with an impressive accuracy of 89.07\%. CABNet followed with a slightly lower accuracy of 85.69\%, while ADCNet performed at 83.40\%. HANet method showed an accuracy of 85.54\%, which underperforms the accuracy achieved by the proposed ensemble model. This highlights the effectiveness of the proposed ensemble method in achieving the highest classification accuracy among the methods tested.
\begin{table}[h]
\centering
\caption{Comparison of Proposed Ensemble Model with Existing Methods on APTOS 19 Dataset}
\begin{tabular}{lll}
\hline
\textbf{References} & \textbf{Methods}            & \textbf{Accuracy} \\ \hline
Chollet et al. \cite{chollet2017xception}  & Xception                    & 86.40\%          \\ 
He et al. \cite{he2020cabnet}       & CABNet                      & 85.69\%          \\ 
Yue et al. \cite{yue2023attention}      & ADCNet                      & 83.40\%          \\ 
Shaik et al. \cite{shaik2022hinge}    & HANet                       & 85.54\%          \\ 
This study          & Proposed Optimized Ensemble & 89.07\%          \\ \hline
\end{tabular}
\end{table}
\subsection{Additional test}
The DDR dataset \cite{wang2020eca} contains a total of 13,673 fundus images. These images are categorized into five groups based on international standards. Images of poor quality or unclear visibility of lesions are placed in a separate class labeled as ungraded. For our experiments, we exclude the ungraded images from the dataset. After removing these ungraded images, we are left with 6,320 images for training, 2,503 images for validation, and 3,759 images for testing.

The performance of the proposed ensemble model is compared to three base models on the additional DDR dataset by using key evaluation metrics. As outlined in Table 7, DenseNet169 achieves a precision of 74.13\%, recall of 71.37\%, F1-score of 71.71\%, and overall accuracy of 81.14\%. MobileNetV1 performs slightly better in terms of recall by reaching 77.22\% but has a lower precision of 73.69\% compared to DenseNet169 model. The F1-score for MobileNetV1 is 78.56\%, and its overall accuracy is 81.69\%. Xception outperforms both DenseNet169 and MobileNetV1 by achieving a precision of 79.62\%, recall of 81.92\%, F1-score of 82.13\%, and overall accuracy of 82.22\%. However, the proposed ensemble model demonstrates superior performance, with a precision of 80.02\%, recall of 83.99\%, F1-score of 84.34\%, and accuracy of 84.35\%. This shows that the proposed ensemble model significantly improves the balance between precision and recall by leading to higher overall accuracy compared to the base models, thus demonstrating its effectiveness in enhancing the detection performance for DR.
\begin{table}[h]
\centering
\caption{Performance comparison among base models and the proposed ensemble model on additional DDR dataset}
\begin{tabular}{lcccc}
\hline
\textbf{Models} & \textbf{Precision (\%)} & \textbf{Recall (\%)} & \textbf{F1-Score (\%)} & \textbf{Accuracy (\%)} \\
\hline
DenseNet169 (Base 1) & 74.13 & 71.37 & 71.71 & 81.14 \\

MobileNetV1 (Base 2) & 73.69 & 77.22 & 78.56 & 81.69 \\

Xception (Base 3) & 79.62 & 81.92 & 82.13 & 82.22 \\
Proposed Ensemble Model & 80.02 & 83.99 & 84.34 & 84.35 \\
\hline
\end{tabular}
\end{table}
Fig 8 illustrates a comprehensive comparison of the CM for the base models and the proposed ensemble model on additional DDR dataset. Among the base models, DenseNet169 (Base 1) in Fig 8(a) correctly classified 3050 out of 3759 test images, resulting in 709 misclassifications. While this indicates relatively high performance, there is still potential for further reduction in misclassifications. MobileNetV1 (Base 2) in Fig 8(b) showed a slight improvement by correctly classifying 3071 out of 3759 images, leaving 688 misclassifications. Xception (Base 3) in Fig 8(c) performed similarly, correctly classifying 3091 out of 3759 test samples, with 668 misclassifications. In comparison, the proposed ensemble model in Fig 8(d) demonstrated a significant enhancement in performance. It correctly classified 3171 out of 3759 test images, reducing the number of misclassifications to just 588. This reduction in misclassifications across all classes underscores the effectiveness of the ensemble model. By leveraging the strengths of each base model and incorporating advanced techniques such as SSA and image fusion, the ensemble model effectively addresses the limitations of individual models, resulting in a substantial improvement in overall classification accuracy and a decrease in misclassification rates.
\begin{figure}[h]
\centering
\begin{minipage}[]{6cm}
  \centering
  \includegraphics[width = 5cm]{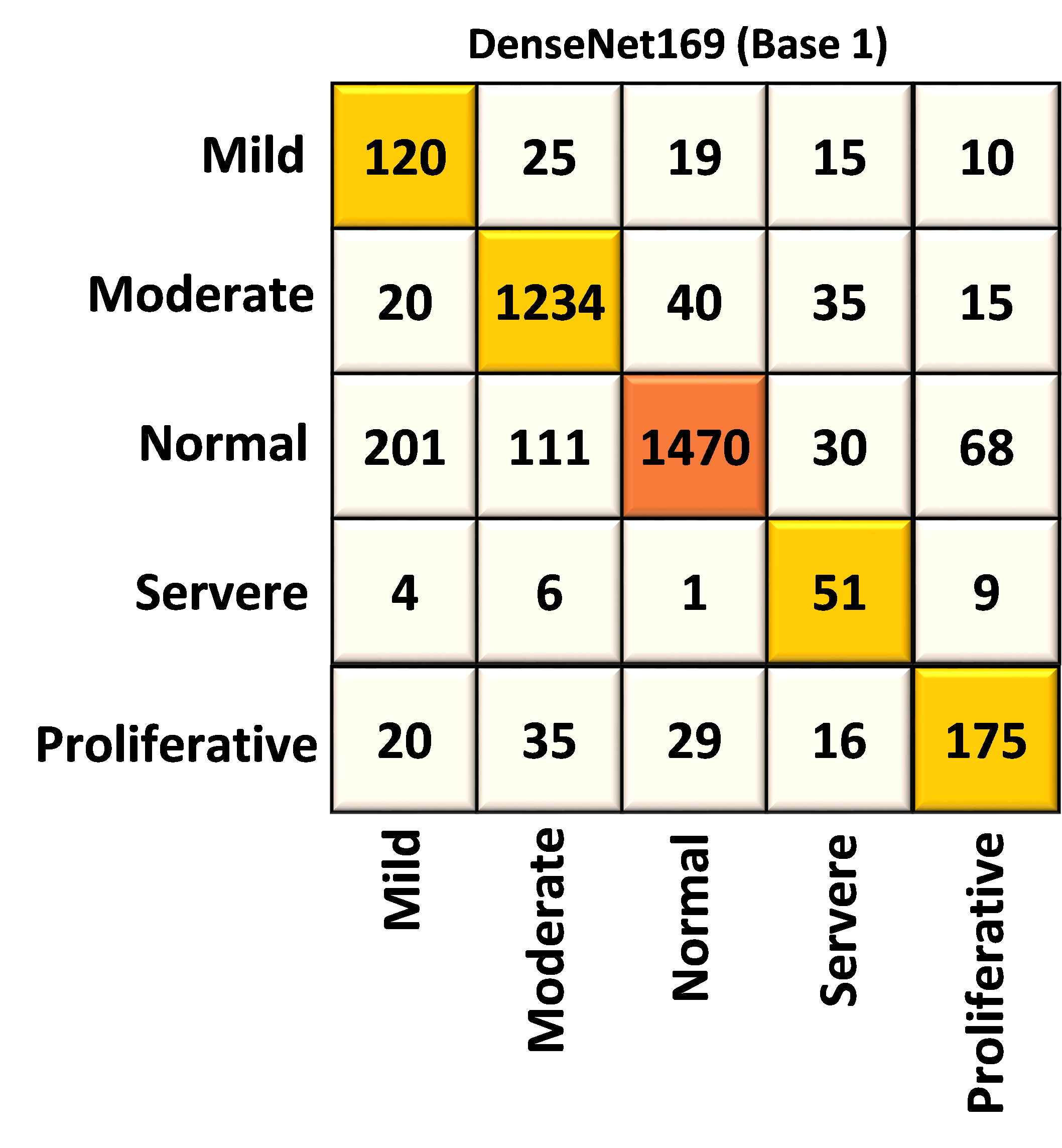}
  \vspace{-0.5cm} 
    \begin{center}
    \textbf{(a)}    
    \end{center}
\end{minipage}
\begin{minipage}[]{6cm}
  \centering
  \includegraphics[width = 5cm]{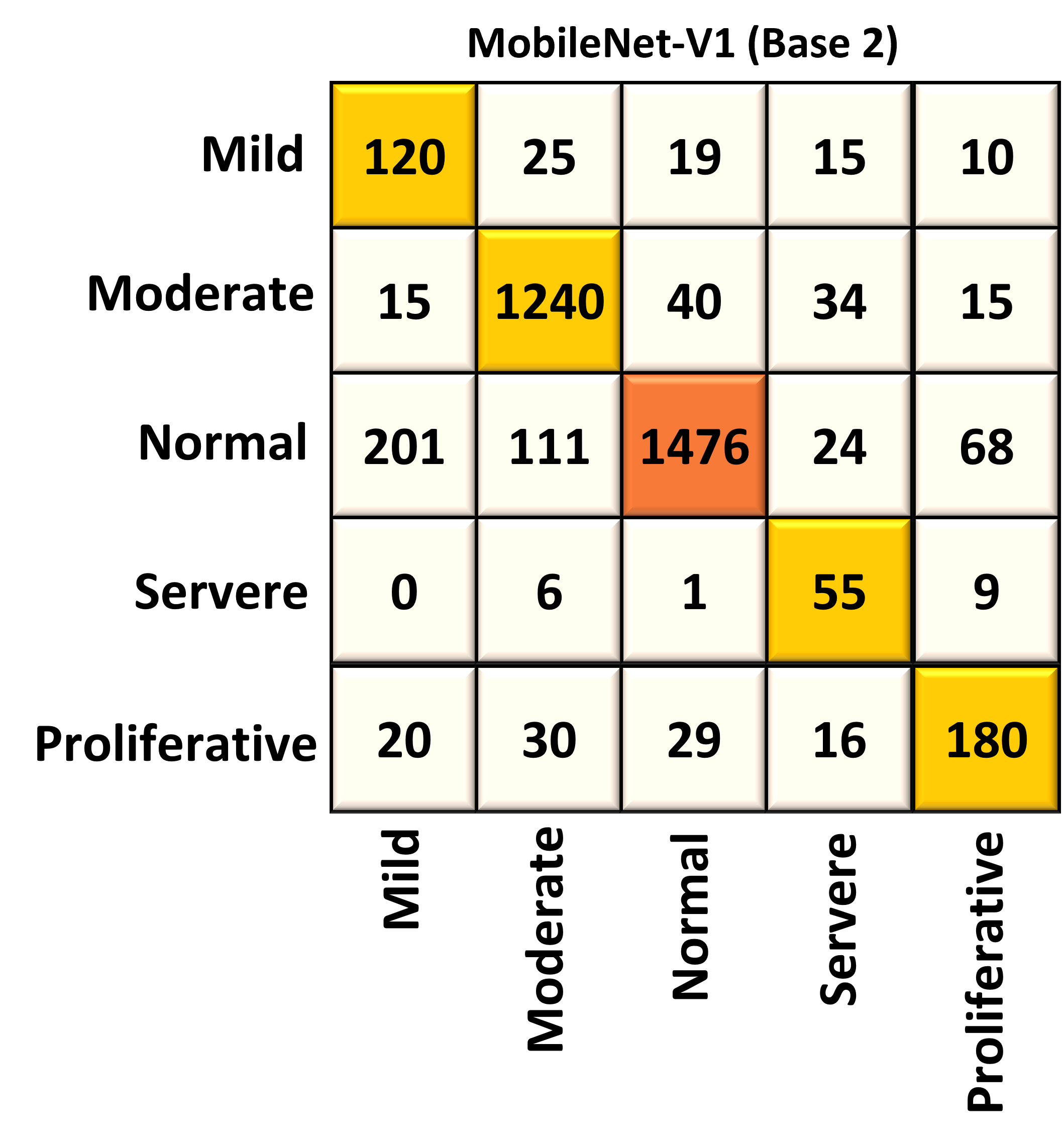}
  \vspace{-0.5cm} 
    \begin{center}
    \textbf{(b)}    
    \end{center}
\end{minipage}
  \begin{minipage}{6cm}
    \centering
    \includegraphics[width = 5cm]{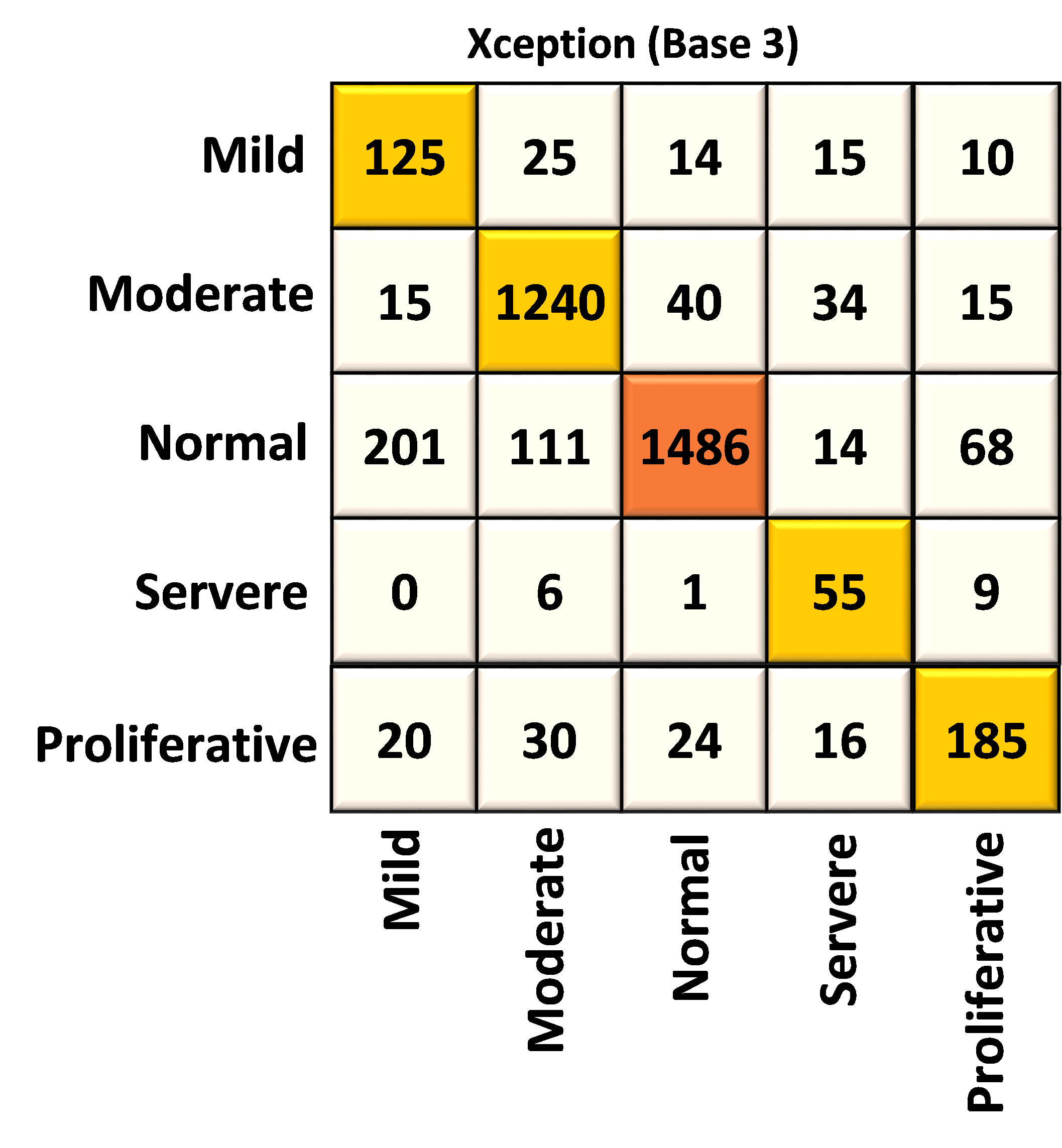}
    \vspace{-0.5cm} 
    \begin{center}
    \textbf{(c)}    
    \end{center}
  \end{minipage}
  \begin{minipage}{6cm}
    \centering
    \includegraphics[width = 5cm]{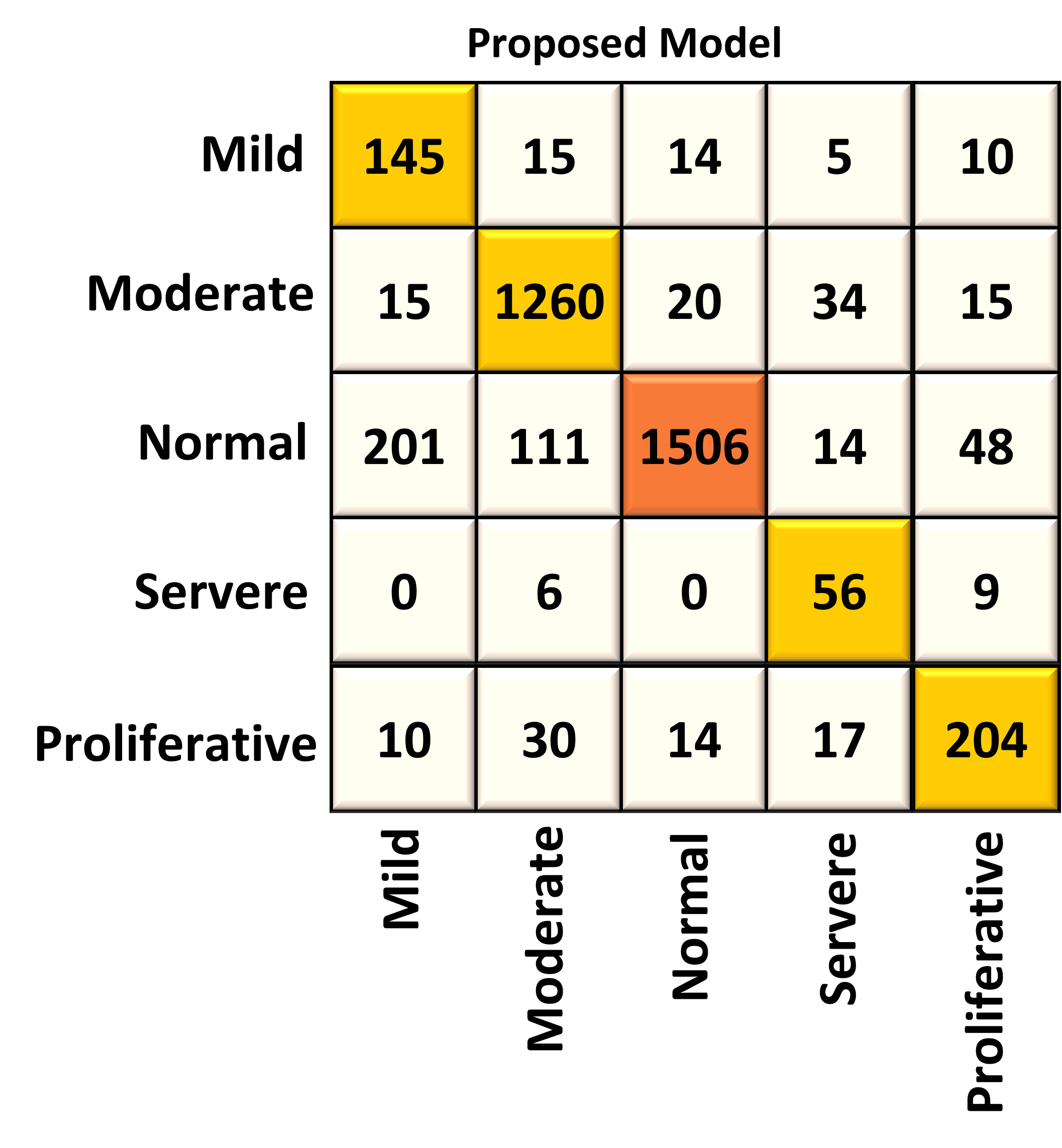}
    \vspace{-0.5cm} 
    \begin{center}
    \textbf{(d)}    
    \end{center}
  \end{minipage}
  \caption{Performance evaluation using CM for (a) DenseNet169 (Base 1), (b) MobileNetV1 (Base 2), (c) Xception (Base 3), and (d) the proposed ensemble model on the additional DDR dataset.}
\end{figure}
\subsection{Ablation study}
We also conducted experiments on the three selected base models with various blocks such as convolutional block attention module(CBAM), channel attention(CA), squeeze and excitation (SE), and residual block(RB), Inception block(IB), dense block(DB) and naive block(NB) , to compare their performance with the chosen improved Inception block. The performance analysis of these models evaluated on the Kaggle APTOS 2019 dataset is presented in Table 9. The performance of models with these blocks is evaluated based on key performance metrics. The DenseNet169-Base + RB model delivers the highest performance, achieving an accuracy of 86.06\% by demonstrating a significant improvement over the other variations. DenseNet169-Base + CBAM achieves an accuracy of 80.87\%, while DenseNet169-Base with SE and CA shows similar accuracy of 80.60\% but still falls behind the performance of the RB-enhanced model. DenseNet169-Base with IB and DB exhibit almost similar accuracy performance with values of 80.33\% and 80.32\%, respectively. DenseNet169-Base + NB performs the second best after the proposed RB by achieving an accuracy of 81.42\%. DenseNet169-Base + IB shows the lowest recall value at 70.10\%, while DenseNet169-Base + RB achieves the highest recall score of 82.51\%. All other blocks show lower recall scores compared to the RB-enhanced model, with values ranging from 70.25\% to 78.02\%.
  
The MobileNetV1-Base + RB model delivers the highest performance, achieving an accuracy of 85.51\% by demonstrating a significant improvement over the other tested blocks. MobileNetV1-Base + CBAM achieves the second-best accuracy of 79.78\% after the proposed RB, while MobileNetV1-Base with SE,IB and NB shows similar accuracy at 79.50\%, but still falls behind the performance of the RB-enhanced model. MobileNetV1-Base with CA and DB exhibit accuracy values of 78.96\% and 78.68\%, respectively. All blocks show lower performance compared to selected RB block. In terms of recall, MobileNetV1-Base + RB also performs the best by showing a recall score of 80.60\%, while MobileNetV1-Base + IB shows the lowest recall value at 65.80\%. The RB block significantly enhances recall, with the IB block performing the least effectively in comparison. The recall values for the other blocks fell between 68.16\% and 74.90\%, which is lower than the recall rate of selected RB-enhanced model. Therefore, the RB block significantly enhances both accuracy and recall by outperforming all other blocks in terms of overall performance. The Xception-Base + RB model delivers the highest performance by demonstrating an accuracy of 85.24\%, significantly surpassing the other blocks. The Xception-Base with DB and NB exhibit similar accuracy of 81.69\% which is second highest accuracy after the selected RB-enhanced model. Other blocks enhanced models exhibiting accuracies in range between 78.68\% and 80.05\% respectively, though they still lagging behind the RB-enhanced model. In terms of recall, Xception-Base + CBAM shows the highest recall value at 79.34\%, while Xception-Base + CA shows the lowest recall at 73.69\%. The recall values for the other blocks fell between 75.92\% and 77.49\%. Although the CBAM block has the highest recall rate, its overall accuracy is lower compared to the RB block. The RB block shows slightly lower recall but achieves higher accuracy by making it a better choice overall among the different blocks. This balance of higher accuracy and slightly lower recall makes the RB block more effective for our model.
\begin{table}[h]
\centering
\caption{Performance Analysis for Baseline Model with Residual Learning and Other Existing Blocks: Kaggle APTOS 2019}
\begin{tabular}{lcccc}
\hline
\textbf{Models}                       & \textbf{Precision (\%)} & \textbf{Recall (\%)} & \textbf{F1-Score (\%)} & \textbf{Accuracy (\%)} \\ \hline
DenseNet169-Base + CBAM                & 75.06                  & 77.56                & 69.53                  & 80.87                 \\ 
DenseNet169-Base + CA                  & 71.77                  & 78.02                & 89.20                  & 80.60                 \\ 
DenseNet169-Base + SE                  & 82.08                  & 77.19                & 88.53                  & 80.60                 \\ 
DenseNet169-Base + RB                  & 83.83                  & 82.51                & 82.85                  & 86.06                 \\ 
DenseNet169-Base + IB                  & 73.44                  & 70.10                & 70.25                  & 80.33                 \\ 
DenseNet169-Base + DB                  & 71.47                  & 70.25                & 70.15                  & 80.32                 \\ 
DenseNet169-Base + NB                  & 74.66                  & 72.09                & 73.17                  & 81.42                 \\ 
MobileNetV1-Base + CBAM                & 60.05                  & 74.44                & 75.68                  & 79.78                 \\ 
MobileNetV1-Base + CA                  & 73.18                  & 68.16                & 68.27                  & 78.96                 \\ 
MobileNetV1-Base + SE                  & 70.92                  & 68.21                & 67.81                  & 79.50                 \\ 
MobileNetV1-Base + RB                  & 83.20                  & 80.60                & 80.78                  & 85.51                 \\ 
MobileNetV1-Base + IB                  & 67.51                  & 65.80                & 66.26                  & 79.50                 \\ 
MobileNetV1-Base + DB                  & 71.80                  & 69.55                & 60.64                  & 78.68                 \\ 
MobileNetV1-Base + NB                  & 79.25                  & 74.90                & 76.29                  & 79.50                 \\ 
Xception-Base + CBAM                   & 71.89                  & 79.34                & 78.91                  & 78.96                 \\ 
Xception-Base + CA                     & 70.53                  & 73.69                & 75.06                  & 79.78                 \\ 
Xception-Base + SE                     & 74.82                  & 79.05                & 79.55                  & 78.68                 \\ 
Xception-Base + RB                     & 81.11                  & 77.49                & 80.29                  & 85.24                 \\ 
Xception-Base + IB                     & 79.62                  & 75.92                & 77.13                  & 80.05                 \\ 
Xception-Base + DB                     & 76.32                  & 78.18                & 70.34                  & 81.69                 \\ 
Xception-Base + NB                     & 73.69                  & 77.22                & 78.56                  & 81.69                 \\ \hline
\end{tabular}
\end{table}
\subsection{Limitations and Advantages}
Due to the deep architecture and high computational performance of the model, pruning techniques need to be considered to optimize efficiency and reduce computational overhead. Additionally, the class imbalance problem persists, which could negatively impact the model's performance, particularly on minority classes. While the model performs well overall, it still faces challenges with misclassification, especially when dealing with edge cases or unseen data, which highlights the need for further improvements in generalization and robustness. However, despite these limitations, our proposed model shows significant promise and serves as a strong foundation for future advancements and refinements.
The key outcome of this study as follow:
\begin{itemize}
\item The integration of image pre-processing techniques, such as CLAHE, Gamma correction and image fusion also strengthens the ability of proposed model to detect DR more efficiently. This process improves the robustness and reliability of the model on different retinal images.
\item The proposed ensemble model improves classification accuracy by combining multiple pre-trained models. This approach enhances feature extraction capabilities and improves overall performance.
\item By using SSA for weight optimization, the model is able to dynamically adjust the contribution of each base model to ensure that the prediction ability of the ensemble model is maximized.
\end{itemize}
\section{Conclusion}\label{sec6}
Diabetic retinopathy is a significant global health issue that can lead to blindness if not diagnosed early. Early detection is crucial for preventing severe vision loss and improving patient outcomes. In this research, we developed a novel ensemble model for DR detection by combining advanced image processing techniques and optimization strategies to enhance model performance. We integrated three pre-trained base models like DenseNet169, MobileNetV1, and Xception to optimize their contribution using the Salp Swarm Algorithm to dynamically assign weights. By leveraging image pre-processing methods such as CLAHE and Gamma correction, followed by image fusion with Discrete Wavelet Transform (DWT), we created a more informative dataset that significantly improves feature extraction for DR detection. The SSA optimization allowed us to intelligently allocate the weight distribution by addressing the limitations of traditional ensemble methods. Through rigorous evaluation on the multiclass Kaggle APTOS 2019 dataset, our proposed model achieved an impressive accuracy of 89.07\% by demonstrating its robustness and effectiveness in DR detection. Furthermore, we assessed model performance with various metrics such as P-R curve, confusion matrix, and ROC curve, plots, all of which confirmed the superior accuracy and transparency of proposed model in decision-making. The contributions of this work include the development of a dynamic ensemble model for DR detection, advanced feature extraction through image fusion, and optimization of model performance using SSA. 

Future work will focus on developing a real-time mobile application for DR detection to improve accessibility for clinicians and patients. We will also explore advanced fusion techniques and unsupervised learning to enhance feature extraction and classification capabilities. The search space will be further optimized by expanding the dataset and introducing more efficient meta-heuristic algorithms. In addition, we plan to address the class imbalance problem to reduce the occurrence of misclassification.
\section*{Declarations}
\bmhead{Declaration of competing interest}
The authors declare that they have no known competing financial interests or personal relationships that could have appeared to influence the work reported in this paper.
\bmhead{Data availability}
Data will be made available on request.
\bmhead{CRediT authorship contribution statement}
Saif Ur Rehman Khan \& Muhammed Nabeel Asim: Conceptualization, Data curation, Methodology, Software, Validation, Writing original draft \& Formal analysis. Sebastian Vollmer: Conceptualization, Funding acquisition, Supervision. Andreas Dengel: review \& editing. 

\bibliography{sn-bibliography}

\end{document}